\documentclass[lettersize,journal]{IEEEtran}
\usepackage{amsmath,amsfonts}
\usepackage{algorithmic}
\usepackage{algorithm}
\usepackage{array}
\usepackage[caption=false,font=normalsize,labelfont=sf,textfont=sf]{subfig}
\usepackage{textcomp}
\usepackage{stfloats}
\usepackage{url}
\usepackage{verbatim}
\usepackage{graphicx}
\usepackage{xcolor}
\usepackage{amsmath,amssymb,amsopn,amstext,amsfonts}
\usepackage[space]{cite}
\usepackage{hyperref}
\hypersetup{
    colorlinks=true,
	citecolor=blue,
	linkcolor=blue,
	urlcolor=blue
}
\usepackage{url}
\usepackage{colortbl}
\usepackage{booktabs}
\usepackage{graphicx}
\usepackage{amsmath}
\usepackage{bm}
\usepackage{makecell}
\usepackage{amssymb}
\usepackage{graphicx,wrapfig,lipsum}
\usepackage{caption}
\usepackage{multirow}
\usepackage{amssymb}
\usepackage{pifont}
\usepackage{cuted}
\usepackage{amssymb}%
\usepackage{pifont}%
\newcommand{\cmark}{\ding{51}}%
\newcommand{\xmark}{\ding{55}}%

\newcommand{\bC}{\mathbf{C}}
\newcommand{\bD}{\mathbf{D}}

\newcommand{\bI}{\mathbf{I}}

\newcommand{\bK}{\mathbf{K}}

\newcommand{\bM}{\mathbf{M}}
\newcommand{\bn}{\mathbf{n}}
\newcommand{\bO}{\mathbf{O}}
\newcommand{\bp}{\mathbf{p}}\newcommand{\bP}{\mathbf{P}}

\newcommand{\bS}{\mathbf{S}}

\newcommand{\bu}{\mathbf{u}}
\newcommand{\bV}{\mathbf{V}}
\newcommand{\bw}{\mathbf{w}}
\newcommand{\bX}{\mathbf{X}}

\newcommand{\bz}{\mathbf{z}}

\makeatletter
\DeclareRobustCommand\onedot{\futurelet\@let@token\@onedot}
\def\@onedot{\ifx\@let@token.\else.\null\fi\xspace}

\makeatother

\usepackage{amsmath}

\DeclareMathOperator*{\argmin}{arg\,min}
\newcommand{\change}[1]{\textcolor{black}{#1}}
\newcommand{\changetwo}[1]{\textcolor{black}{#1}}

\renewcommand{\hm}[1]{{\color{black}#1}}
\begin{document}

\title{SurgTAP-3D: A Large-Scale Cross-Surgery Video Dataset with Contact-Aware 3D Motion Tracking}
\author{Jiaxin Guo,~\IEEEmembership{Member,~IEEE}, Wenzhen Dong,~\IEEEmembership{Student Member,~IEEE}, Taoyu Wu,~\IEEEmembership{Student Member,~IEEE}, \\Haomin Kuang, Mengya Xu, Kejian Shi,~\IEEEmembership{Student Member,~IEEE}, Ziyi Wang,~\IEEEmembership{Member,~IEEE}, \\Tongfan Guan, Mingkai Jia,~\IEEEmembership{Student Member,~IEEE}, Yue Wang,~\IEEEmembership{Senior Member,~IEEE}, \\Qi Dou,~\IEEEmembership{Senior Member,~IEEE}, Yun-Hui Liu$^\dagger$,~\IEEEmembership{Fellow,~IEEE}
\thanks{This work is supported in part by the HK RGC AoE under AoE/E-407/24-N, in part by RGC GRF Ref. 14207324, in part by the InnoHK initiative of the Innovation and Technology Commission of the Hong Kong Special Administrative Region Government via the Hong Kong Centre for Logistics Robotics, in part by the Multi-Scale Medical Robotics Centre, and in part by the CUHK T Stone Robotics Institute. (\textit{Corresponding author: Yun-Hui Liu})}%
\thanks{Jiaxin Guo, Wenzhen Dong, Ziyi Wang, Tongfan Guan, Qi Dou, and Yun-Hui Liu are with the T Stone Robotics Institute and Hong Kong Centre for Logistics Robotics, The Chinese University of Hong Kong, Hong Kong SAR, China. Taoyu Wu is with the University of Liverpool, Liverpool, U.K. Haomin Kuang is with the School of Mechanical Engineering, Shanghai Jiao Tong University, Shanghai, China. Mengya Xu, Kejian Shi, and Qi Dou are with the Department of Computer Science and Engineering, The Chinese University of Hong Kong, Hong Kong SAR, China. Mingkai Jia is with the Department of Electronic and Computer Engineering, The Hong Kong University of Science and Technology, Hong Kong SAR, China.
Yue Wang is with the College of Control Science and Engineering, Zhejiang University, Hangzhou, China.}

}

\maketitle

\begin{strip}
\vspace{-1.7in}
\setlength\tabcolsep{0.5pt}
\centering

\includegraphics[width=1.0\linewidth]{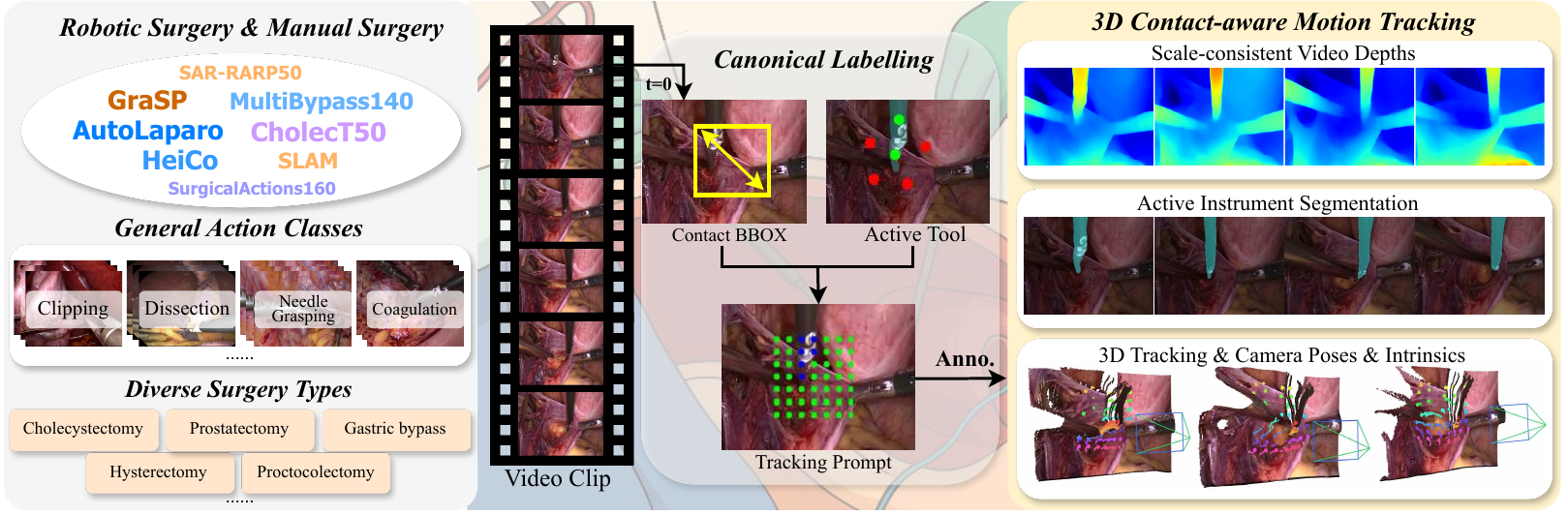}
\captionof{figure}{\textbf{\hm{We present SurgTAP-3D as the first large-scale 3D motion tracking dataset from dynamic real-world surgery videos.}} Our dataset encompasses 4.5K sequences with 520K frames from diverse real-world robotic and manual surgeries, collecting the clips under unified action classes. Through a simple canonical labelling, \change{our proposed framework enables reliable and accurate 3D reconstruction and tracking}, producing pseudo-scale pointclouds, 3D point trajectories of instruments and tissues, camera poses, and active tool segmentations.}
\label{fig: teaser_title}
\end{strip}

\begin{abstract}
Understanding 3D scene dynamics in minimally invasive surgery (MIS) is a fundamental step toward enabling physically intelligent and autonomous surgical systems. However, existing public datasets are largely confined to ex-vivo or simulated environments, offering limited scalability and diversity in real-world clinical scenarios.  Moreover, current annotations typically focus on either tissue or instruments alone, overlooking the critical contact dynamics that govern tool–tissue interactions. This gap limits the downstream applications such as augmented reality guidance and autonomous surgery driven by vision-language-action (VLA) models or world models, which demand large-scale, high-quality 3D supervision. 
\change{In this work, we construct SurgTAP-3D, a large-scale in-vivo pseudo-3D surgical motion tracking dataset and benchmark, comprising 520K frames, 4,549 action clips, and 1.4M 2D/3D point trajectories from diverse robotic and manual procedures under unified action categories. To make such annotation scalable, we design the first unified framework for Surgical 3D Point Tracking that converts uncalibrated monocular surgical videos into structured 4D annotations through contact-region prompting, active-tool segmentation, dynamic bundle adjustment, and global 4D refinement. This pipeline adapts foundation priors to surgical videos and produces temporally consistent trajectories, pseudo-scale geometry, camera poses, and instrument masks.}
The dataset captures realistic challenges inherent to surgical environments, including occlusions, smoke, bleeding, and rapid instrument motion. Furthermore, we establish a comprehensive benchmark to evaluate 3D point tracking performance across the SurgPose, SurgT, and STIR datasets, covering both instrument and tissue tracking scenarios. We also demonstrate downstream applications such as instrument displacement estimation, deformable surface reconstruction, and fine-tuning foundation models, highlighting the practical utility of our dataset and framework. Extensive experiments and analysis show the strong robustness and effectiveness \change{of our dataset and annotation framework, for 3D surgical point tracking, reconstruction applications, and surgical action understanding, addressing an important limitation in real-world 3D surgical motion data. Homepage: \url{https://wrld.github.io/surgtap-3d/}}.
\end{abstract}


\section{Introduction}
\label{sec:intro}
        
\begin{table*}[t]
    \small
    \centering
    \caption{\textbf{Comparison with Surgical Tracking Datasets.} 
    Our \textbf{SurgTAP-3D} is the first large-scale, \textbf{in-vivo}, \textbf{contact-aware}, \textbf{3D pixel-wise trajectory} dataset covering \textbf{both instrument and tissue} in real robotic and manual surgeries. 
    \textcolor{gray}{\textbf{Gray row}}: ours.}
    \setlength{\tabcolsep}{0.5mm}
    \renewcommand{\arraystretch}{1.15}
    \begin{tabular}{l c c c c c c c c c c c}
        \toprule
        \multirow{2}{*}{\textbf{Dataset}} & 
        \multirow{2}{*}{\makecell{\textbf{Surgical}\\\textbf{Domain}}} & 
        \multirow{2}{*}{\textbf{Setup}} & 
        \multirow{2}{*}{\textbf{In-vivo}} & 
        \multirow{2}{*}{\makecell{\textbf{Contact}\\\textbf{Aware}}} &
        \multirow{2}{*}{\makecell{\textbf{Annot.}\\\textbf{Object}}} & 
        \multicolumn{3}{c}{\textbf{Annotation Type}} & 
        \multicolumn{3}{c}{\textbf{Dataset Scale}} \\
        \cmidrule(lr){7-9} \cmidrule(lr){10-12}
        & & & & & &\textbf{2D}& \textbf{3D} & \textbf{Pixel Traj.}  & \textbf{Seqs} & \textbf{Trajs} & \textbf{Frames} \\
        \midrule
        SurgT~\cite{cartucho2024surgt}          & Multiple Procedures      & \makecell{Manual, \\ Robotic}   & Partial & \xmark & Tissue&\cmark & \cmark   & \xmark  & 157  & 32       & 24{,}548   \\
        STIR~\cite{schmidt2025point}             & \makecell{Cholecystectomy, \\Nissen Fundoplication Surgery}      & Robotic  & Partial & \xmark & Tissue&\cmark & \cmark   &  \cmark  & 576  & 496 & 13{,}393         \\
        CholecTrack20~\cite{nwoye2025cholectrack20} & Cholecystectomy Surgery  & Manual  & \cmark & \xmark & Instrument&\cmark & \xmark   &  \xmark   & 20   & 2{,}624 & 35{,}000  \\
        CRCD~\cite{oh2024comprehensive}         & Ex-vivo pseudo-cholecystectomy        & Robotic  & \xmark & \xmark & Instrument&\cmark & \xmark   & \xmark & 16   & -       & 127{,}000 \\
        SurgPose~\cite{wu2025surgpose}          & Instrument Motion over Tissue   & Robotic  & \xmark & \xmark & Instrument&\cmark & \cmark   & \cmark & 30   & 476 & 30{,}030  \\
        SurgRIPE~\cite{xu2025surgripe}          & Instrument Manipulation    & Robotic  & \xmark & \xmark & Instrument&\cmark & \cmark   & \xmark & 32    & -       & 2,841         \\
        \change{StereoMIS~\cite{hayoz2024}} & \change{\makecell{Porcine Endoscopic Surgery}} 
        & \change{Robotic} 
        & \change{\cmark} 
        & \change{\xmark} 
        & \change{Tissue} 
        & \change{\cmark} 
        & \change{\xmark} 
        & \change{\cmark} 
        & \change{7} 
        & \change{21--28} 
        & \change{1{,}400} \\
        \midrule
        \rowcolor[gray]{0.9}
        \textbf{SurgTAP-3D}               & \textbf{\makecell{Cholecystectomy, Prostatectomy, \\Gastric bypass, Hysterectomy, \\Proctocolectomy, etc.}} & \textbf{\makecell{Manual, \\ Robotic}} & \cmark & \cmark& \textbf{\makecell{Tissue, \\ Instrument}} & \cmark & \cmark & \cmark& \textbf{4{,}549} & \textbf{1{,}376{,}987} & \textbf{519{,}642} \\
        \bottomrule

    \end{tabular}
    \label{tab:benchmark_comparison}
    \vspace{-1.2em}
\end{table*}

Intelligent and autonomous systems are rapidly emerging as a cornerstone of next-generation surgical robotics, enabling advanced capabilities such as autonomous soft-tissue manipulation~\cite{hu2024occlusion}, real-time intraoperative navigation~\cite{liu2025aisearch}, and augmented reality–assisted interventions~\cite{mccloskey2023vrar}.
These technologies enhance situational awareness, decision support, and execution accuracy during complex procedures.
Recent advances in data-driven and multimodal learning have further accelerated progress toward surgical autonomy~\cite{zeng2026surg}, including language-conditioned imitation learning transformers \cite{kim2025srt}, vision–language–action (VLA) models for end-to-end suturing \cite{haworth2025suturebot}, and world models for visuomotor policy learning \cite{lin2025visuomotor,xu2026generalized}. 
These advances demonstrate the potential of learning-based surgical automation to improve execution precision, reduce surgeon workload, and enhance procedural safety.

Despite these advances, substantial challenges still impede the realization of fully autonomous, long-horizon surgical tasks.
A primary limitation is the scarcity of large-scale, high-fidelity endoscopic datasets with synchronized visual observations and robotic kinematics, which constrains the scalability and generalization of policy learning frameworks.
Existing surgical datasets with synchronized video and robotic kinematics remain limited in scale and task diversity~\cite {haworth2025suturebot,wu2025surgpose,oh2024comprehensive}.
For instance, SRT-H \cite{kim2025srt} proposes a hierarchical framework for dexterous and long-horizon surgical automation but relies on trajectories collected from 31 porcine gallbladders, restricting task diversity and limiting cross-procedure generalization. 
Similarly, SutureBot \cite{haworth2025suturebot} provides 1,890 dVRK demonstrations for long-horizon suturing; however, these data are collected in tissue phantoms, leading to a substantial sim-to-real domain gap.
In contrast, recent advances in VLA-based models~\cite{zitkovich2023rt,kim2024openvla,zhou2025vision} and world models~\cite{cen2025worldvla,team2025gigabrain,jang2025dreamgen} typically rely on datasets comprising thousands to millions of temporally extended interaction sequences in diverse tasks, embodiments, and environments to achieve robust generalization. 
Existing surgical demonstration datasets remain limited in scale and diversity, restricting the generalization of high-capacity multimodal policies. Therefore, advancing surgical automation requires the collection of large-scale, high-quality real-world demonstrations that provide temporally synchronized action kinematics and endoscopic image streams to support robust and generalizable policy learning.

Despite the availability of diverse public datasets in surgical vision research, existing datasets still face several key limitations that hinder the development of scalable and generalizable models for 3D motion tracking in real-world applications, as reported in Tab.~\ref{tab:benchmark_comparison}.
Firstly, most datasets~\cite{oh2024comprehensive, wu2025surgpose, xu2025surgripe} are collected in \textit{ex-vivo} or simulated environments rather than \textit{in-vivo} surgical scenes, where visual appearance, illumination, and tissue dynamics differ substantially from real-world conditions.
Secondly, the scalability and diversity of existing datasets remain limited, as many focus on a single surgical procedure or a constrained acquisition setup (e.g., exclusively manual or robotic). Such restrictions substantially limit cross-procedure variability and hinder generalization across heterogeneous clinical environments and operational conditions.
Finally, annotated targets are typically restricted to either tissue or instruments, resulting in limited representational flexibility and a lack of \textit{contact-aware} annotations that capture tool--tissue interaction regions. \change{We define ``contact-aware'' annotation as explicitly marking visible tool--tissue interaction regions and using them to initialize trajectories around surgical interaction areas, without assuming force sensing or contact-physics modeling.}
This impedes the development of contact-aware perception and physically grounded motion understanding, both of which are essential for autonomous surgical manipulation.
These limitations highlight the urgent need for \textbf{large-scale, cross-surgery, and contact-aware} 3D datasets to enable robust perception, reasoning, and planning for autonomous surgical systems. Motivated by this, we further ask: \textit{given the massive amount of surgical video data available, can we recover high-quality 3D action trajectories from videos?}

Addressing this problem involves two primary challenges. 
First, in the absence of priors such as kinematics or camera calibration, recovering accurate camera parameters, scene geometry, and instrument motion from monocular surgical videos is inherently difficult~\cite{wei2024scale}. 
Second, contact-aware labeling is labor-intensive, often requiring substantial manual effort~\cite{nwoye2025cholectrack20,cartucho2024surgt} or specialized physical markers, such as tattoo needles used in STIR~\cite{schmidt2025point}.
To address these challenges, we draw inspiration from recent advances in large-scale foundation models for 3D reconstruction and motion tracking, which have demonstrated the ability to learn accurate geometry and motion representations in a feed-forward manner. 
Traditional multi-stage pipelines, such as structure-from-motion (SfM), often suffer from drift accumulation and inconsistent scale. In contrast, modern data-driven reconstruction frameworks~\cite{wang2024dust3r, wang2025vggt, guo2025endo3r,guo2026salon3r} can directly predict camera poses and dense point clouds, achieving scale-consistent depth estimation even in challenging endoscopic environments~\cite{guo2025endo3r}. Similarly, recent 3D motion tracking methods~\cite{xiao2024spatialtracker, xiao2025spatialtrackerv2} and depth foundation models~\cite{guan2025bridgedepth,guan2026dispvit} have shown promise in lifting 2D point tracks to 3D using learned depth priors, paving the way for scalable 3D motion understanding in surgical settings.

Motivated by these insights, we introduce the first unified framework for surgical 3D point tracking, designed to construct large-scale, cross-surgery 3D datasets that facilitate the development of spatially aware autonomous surgical systems.
Our goal is to recover high-fidelity 4D trajectories and 3D geometry from highly dynamic, uncalibrated monocular surgical videos, capturing complex tool-tissue contact in real-world clinical settings.
There are several core designs in our method. 
Firstly, we exploit priors from feed-forward foundation models in reconstruction, point tracking, and segmentation, to form a robust and scalable 4D tracking pipeline. 
Secondly, to mitigate noises and drifting issues from the priors, we employ a \textit{\change{dynamic} bundle adjustment} that explicitly leverages the correspondence field to stabilize and correct long-term drift in camera intrinsics, extrinsics, and dense depth predictions.
Finally, a \textit{global 4D optimization} is introduced to jointly refine geometry and point trajectories by enforcing constraints of instrument rigidity, articulated flow consistency, and temporal smoothness.
These allow SurgTAP-3D to produce spatially coherent, temporally smooth, and physically plausible 3D trajectories of both surgical instruments and deformable tissues, outperforming existing 2D or 3D trackers in surgical settings.

Building upon this framework, we propose \textbf{SurgTAP-3D}, for \textbf{S}urgical \textbf{T}racking \textbf{A}ny \textbf{P}oint in \textbf{3D}, a large-scale, contact-aware 3D motion tracking dataset that provides high-quality and diverse annotations from \textbf{4549 sequences}, \textbf{520K frames}, with \textbf{1.4M 3D trajectories}, including comprehensive annotations for each clip (i.e., intrinsics, extrinsics, depths, 2D and 3D pixel-level trajectories, instrument segmentations).
\change{Our dataset aims to support surgical 3D motion analysis and downstream tasks by providing pseudo-annotated 4D trajectories and scale-consistent scene geometry from uncalibrated surgical videos.}
Specifically, we curate large-scale sequences from a broad spectrum of procedures, including cholecystectomy, prostatectomy, gastric bypass, hysterectomy, proctocolectomy, etc.
Moreover, our datasets are organized into 5 pivotal actions: \textbf{dissection, tissue retraction, needle grasping, vessel clipping, and coagulation}, capturing the most complex tool-tissue interactions.
Unlike prior benchmarks restricted to camera-centric coordinates, \change{SurgTAP-3D provides trajectories in \textbf{both camera and global coordinates}. The global coordinates are obtained by transforming camera-frame 3D points using estimated camera poses, disentangling camera ego-motion from tissue and instrument motion. Under monocular reconstruction, these coordinates remain pseudo-scale unless additional metric calibration is available, supporting 3D motion analysis for surgical perception and manipulation-oriented studies.}
We further build comprehensive benchmarks to assess the current state-of-the-art methods for 3D surgical point tracking during tool-tissue interaction, offering a step toward scalable 3D learning in real-world surgery. Downstream tasks like instrument displacement estimation, deformable surface reconstruction are also conducted and validated to show the applicability.

Our core contributions are summarized as:
\begin{itemize}
\item We build \textbf{SurgTAP-3D}, the first large-scale \textit{in-vivo} 3D motion tracking dataset in general action classes, achieving \textit{contact-aware} annotation from public datasets, encompassing \textbf{520K} frames with \textbf{4549} action clips and \textbf{1.4M} 2D/3D point trajectories in diverse surgeries.
\item We present the first unified framework enabling 3D contact-aware motion tracking to recover high-fidelity 4D trajectories and 3D geometry from dynamic, uncalibrated monocular surgical videos.
    
    \item We build a comprehensive benchmark to assess the current state-of-the-art methods for 2D and 3D point tracking, forming the first 3D tracking benchmark for both instrument and tissue tracking.
    \item Based on our proposed pipeline and dataset, we conduct real-world applications, including instrument displacement estimation, deformable surface reconstruction, \change{demonstrating the applicability to surgical 3D motion analysis and the potential for future robotic applications.}
\end{itemize}

\section{Related Work}
\label{sec:formatting}
\subsection{Surgical Datasets with Motion Annotations}
Surgical datasets play a pivotal role in understanding and modeling scene dynamics for computer-assisted and autonomous surgery. 
However, unlike the diverse and large-scale datasets for motion tracking in general computer vision~\cite{koppula2024tapvid, jin2024stereo4d,doersch2022tap}, only a few surgical datasets provide motion-related annotations such as instrument pose estimation or soft tissue tracking. This lack of large-scale, geometry-aware data significantly limits the development of robust 3D perception, reasoning, and policy learning models for surgical environments.
Regarding the tool motion tracking, SurgPose~\cite{wu2025surgpose} provides instance-aware semantic keypoints and skeletal annotations for surgical tool pose estimation and tracking. CRCD~\cite{oh2024comprehensive} offers an ex-vivo dataset collected on pig livers using the dVRK, with an extended version including keypoint-level annotations of the instruments.
The SurgRIPE Challenge~\cite{xu2025surgripe} releases a dataset for 6D pose estimation of dVRK instrument wrists. However, the data are captured in a controlled laboratory setting rather than from real surgical procedures.
Additionally, CholecTrack20~\cite{nwoye2025cholectrack20}, derived from Cholec80~\cite{twinanda2016endonet} and CholecT50~\cite{nwoye2022rendezvous}, focuses on multi-tool 2D tracking through bounding-box annotations but lacks fine-grained point trajectories and depth-aware 3D tracking.
\change{StereoMIS~\cite{hayoz2024} provides stereo endoscopic sequences with manually annotated landmarks for evaluating 2D tracking in real in-vivo surgical scenes, but its scale is limited and it does not provide 3D tracking annotations.}
SurgT~\cite{cartucho2024surgt} is the first standardized benchmark for soft tissue tracking from Hamlyn~\cite{giannarou2012probabilistic}, SCARED~\cite{allan2021stereo}, and Kidney boundary~\cite{hattab2020kidney} datasets. STIR~\cite{schmidt2025point} is the first point tracking challenge for soft tissue, consisting of stereo video clips from dVRK system and tattoo labels as groundtruth.
Overall, as shown in Tab.~\ref{tab:benchmark_comparison}, current datasets remain limited to ex vivo or simulated settings, lack scale and diversity, and often miss dense 3D motion or contact-level annotations. To bridge this gap, we propose a large-scale, contact-aware 3D motion tracking dataset derived from real surgical videos, offering high-quality annotations for both instrument and tissue motion, advancing scalable and data-driven surgical autonomy.

\subsection{Tracking Any 2D and 3D Point}
Recent advances in video understanding have revolutionized Tracking Any Point (TAP) methods, enabling robust 2D and 3D trajectory estimation for arbitrary pixels across frames. In the 2D domain, TAPIR~\cite{Doersch_2023_ICCV} introduced per-frame initialization with temporal refinement via transformers, achieving state-of-the-art occlusion handling and long-range tracking on benchmarks like TAP-Vid~\cite{doersch2022tap}. CoTracker~\cite{karaev2024cotracker} and its enhanced successor CoTracker3~\cite{karaev2025cotracker3} further improved dense point correspondence by jointly tracking points across sequences using correlation-based attention, reducing tracking errors by up to 40\% on real-world videos with minimal supervision.
Extending to 3D, methods like TAP-Vid-3D~\cite{koppula2024tapvid} provide benchmarks for monocular 3D point tracking, while DELTA~\cite{ngo2024delta} enables efficient dense 3D tracking for entire videos through long-range motion estimation. TAPNext~\cite{zholus2025tapnext} formulates point tracking as next-token prediction using a recurrent architecture for online tracking, boosting accuracy on dynamic scenes. For persistent geometry, TAPIP3D~\cite{zhang2025tapip3d} represents videos as spatio-temporal 3D feature clouds, supporting long-term tracking in RGB-D inputs. Most recently, SpatialTrackerV2~\cite{xiao2025spatialtrackerv2} jointly estimates scene geometry, camera motion, and 3D point trajectories from monocular videos, incorporating bundle adjustment to refine poses.
Despite advances in general and surgical tracking~\cite{guo2024ada,karaoglu2025litetracker,hayoz2024},
specular reflections, tissue deformation, occlusions, and articulated
instrument motion remain challenging.
They often overlook domain-specific priors, such as contact dynamics or semantic segmentation, leading to drift in low-texture regions. To address this, we adopt TAPNext to first obtain the 2D tracking trajectory, and further conduct global 4D optimization to enforce constraints of instrument rigidity, flow consistency, and temporal smoothness, tailoring contact-aware 3D motion tracking in endoscopy.

\subsection{Dynamic Surgical 3D Reconstruction}
Dynamic 3D reconstruction aims to recover the geometry and poses from dynamic videos, which is an ill-posed and under-constrained problem.
Previous deformable methods have been dominated by Shape-from-Template (SfT)~\cite{bartoli2015shape} and Non-Rigid Structure-from-Motion (NRSfM)~\cite{bregler2000recovering}. 
To overcome the static camera limitation, DefSLAM~\cite{lamarca2020defslam} and SD-DefSLAM~\cite{gomez2021sd} are proposed as monocular SLAM frameworks to combine the NRSfM and SfM. 
However, these methods generally rely on isometric assumptions, which restrict their utility in medical applications.
To address this issue, NR-SLAM~\cite{rodriguez2024nr} builds a non-rigid monocular SLAM system based on the dynamic representation composed of a dynamic deformation graph and a visco-elastic deformation model.
However, traditional methods still fall short of reconstructing highly dynamic surgical scenes with dense point clouds, due to their inherent static assumptions and sparse reconstruction.

To address this, recent novel view synthesis techniques have paved the way for dynamic surgical reconstruction~\cite{wang2022neural,zha2023endosurf,yang2024deform3dgs,xie2024surgicalgaussian,li2024endosparse,wu2025endowave,guo2024free}. 
EndoNeRF~\cite{wang2022neural} first explores neural rendering for deformable tissue reconstruction from stereo endoscopic inputs and devises a mask-guided ray-casting strategy to address the tool occlusions. Following this, many NeRF-based approaches~\cite{zha2023endosurf, guo2024uc, yang2023neural, yang2024efficient} are presented to further enhance the performance in rendering, reconstruction quality, and efficiency. 

Deform3DGS~\cite{yang2024deform3dgs} introduces 3DGS with a flexible deformation modeling scheme to learn tissue dynamics for each Gaussian. SurgicalGaussian~\cite{xie2024surgicalgaussian} models the spatial-temporal features by a forward deformation MLP and a regularization for consistent movement. Endosparse~\cite{li2024endosparse} aims to improve the geometric and appearance quality with sparse observations as input.
However, these methods are either restricted to requiring the preprocessed priors from COLMAP~\cite{schonberger2016structure} or requiring per-scene offline optimization, which is time-consuming for real-time applications.

\subsection{Feed-forward 3D Reconstruction}
Recent 3D reconstruction networks also pave the way for joint estimation of geometry in an end-to-end framework~\cite{wang2024dust3r,leroy2024grounding,wang2025continuous,wang2025vggt,guo2025endo3r,guo2026salon3r,zhu2026eps3d}.
This avoids the accumulated errors across decomposed multiple tasks in traditional reconstruction pipelines, such as feature extraction~\cite{detone2018superpoint, yi2016lift}, feature matching~\cite{lindenberger2023lightglue, sarlin2020superglue}, and SfM~\cite{schonberger2016structure, wang2024vggsfm}. To address these limitations, DUSt3R~\cite{wang2024dust3r} proposes a unified framework that reformulates stereo reconstruction as a pointmap regression task, eliminating the reliance on explicit camera calibration. MASt3R~\cite{leroy2024grounding} further enhances the matching accuracy by introducing a dense feature prediction head trained with an additional matching supervision loss.
Feed-forward reconstruction methods are prone to long-term drift, which can accumulate over time and degrade geometric consistency and tracking accuracy. 
To mitigate this issue, we leverage the priors provided by the feed-forward network as initialization, while introducing a \change{dynamic} bundle adjustment module to correct drift and enforce temporal stability. Building upon this, we formulate a global 4D optimization that jointly refines scene geometry and point trajectories, ensuring multi-view consistency and temporally smooth 3D motion predictions across the entire sequence.

\section{SurgTAP-3D}

\begin{figure*}[t]
\centering
\includegraphics[width=1.0\linewidth]{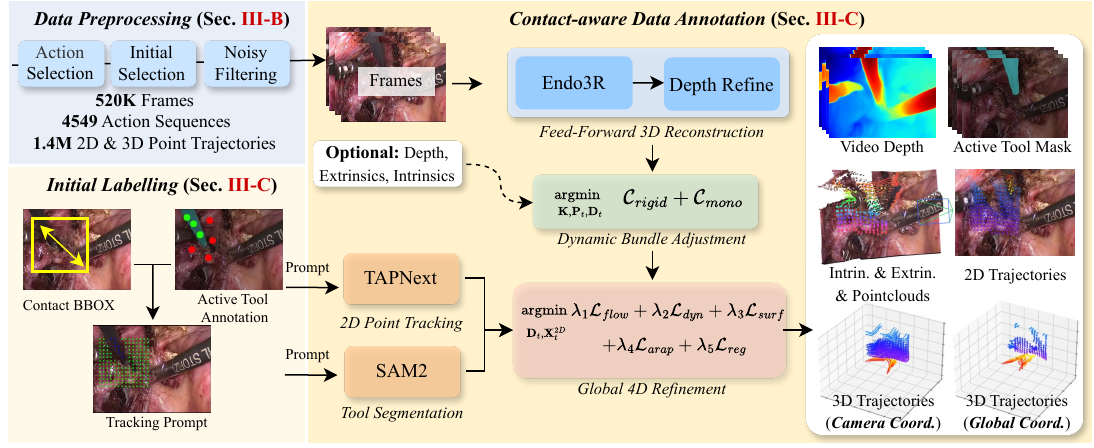}
\caption{\change{\textbf{Overview of the Annotation Framework for SurgTAP-3D.} We first preprocess the data clips from the public datasets to collect the action clips, select the first frame, and filter the noisy frames. The annotation pipeline begins with the canonical labeling for the first frame of the clip, localizing the contact region and active tool to obtain the tracking prompts, which are input into TAPNext and SAM2 for inference. Next, our proposed pipeline takes frames and optional priors as input, achieving stable and robust 3D reconstruction through surgical-aware geometry priors from Endo3R~\cite{guo2025endo3r} and \change{dynamic} bundle adjustment for correcting the drift. Finally, we conduct global 4D refinement to optimize the estimated depths and tracking result, enabling temporally smooth and geometrically coherent 3D contact-aware point tracking.}}
\label{fig: pipeline}
\end{figure*}

In this work, we aim to achieve 3D contact-aware motion tracking, to recover high-fidelity 4D trajectories and 3D geometry from uncalibrated monocular surgical videos. Building upon this, we construct a large-scale dataset of contact-aware surgical clips collected from diverse publicly available surgical videos. Each clip is automatically annotated with rich motion, geometric, and semantic information through our proposed pipeline.
Formally, given a dataset $\mathcal{D} = \{ C_i \}_{i=0}^{Q-1}$ containing $Q$ video clips, each clip $C_i$ is associated with an action label $a_i$, a manually annotated contact region $\mathbf{B}_i$, and an active-tool prompt used to initialize $N$ grid-sampled query points for tracking.
For each clip consisting of $T$ frames, the goal of the annotation framework is to generate comprehensive geometric, motion, and semantic annotations, including the intrinsic camera matrix $\mathbf{K}$, camera poses $\mathbf{P}_t$, depth maps $\mathbf{D}_t$, active-tool segmentations $\mathbf{M}_t^{seg}$, 2D point trajectories $\mathbf{X}_t^{2D}$, corresponding 3D trajectories $\mathbf{X}_t^{3D}$, and \change{per-point visibility estimates $\mathbf{V}_t$}.

As illustrated in Fig.~\ref{fig: pipeline}, the pipeline for SurgTAP-3D is composed of three main stages: input data preprocessing, canonical labelling, and contact-aware automatic annotation. In Sec.~\ref{sec: data_source}, we describe the data sources used in SurgTAP-3D, which include diverse in-vivo surgical videos covering both manual and robotic procedures, organized under a unified set of action categories.  
Sec.~\ref{sec: data_preprocess} presents the data preprocessing process, including frame extraction, quality filtering, and sequence organization.  
Sec.~\ref{sec: pipeline} details the annotation pipeline, which generates spatial and temporal labels through 3D reconstruction, iterative motion refinement, and active instrument segmentation.  
Finally, we provide dataset statistics and analysis in Sec.~\ref{sec: data_stat}.

\subsection{Dataset Source}\label{sec: data_source}
\subsubsection{In-vivo Diverse Surgery Types} 
\change{Existing public datasets provide useful but limited coverage for in-vivo 3D surgical tracking. CRCD~\cite{oh2024comprehensive} and SurgPose~\cite{wu2025surgpose} are mainly collected from ex-vivo pseudo-surgery scenarios with relatively controlled dVRK setups, while CholecTrack20~\cite{nwoye2025cholectrack20} contains real cholecystectomy videos but focuses on a single procedure and provides no 3D geometric annotations. To construct SurgTAP-3D, as shown in Tab.~\ref{tab: release_dataset}, we therefore curate in-vivo surgical videos from multiple procedures, including prostatectomy, hysterectomy, cholecystectomy, gastric bypass, rectal resection, etc. For each selected clip, we further generate synchronized annotations including camera intrinsics, camera poses, depth maps, instrument masks, and dense 2D/3D trajectories for both tissue and instruments.}

\subsubsection{Manual and Robotic Surgeries} The dynamics and operational paradigms of robotic and manual surgeries differ substantially. 
Robotic surgeries~\cite{xu2025surgripe} often exhibit smooth, stable, and kinematically constrained tool motions with a fixed camera view, while manual procedures feature freehand endoscope movement and more dynamic tool trajectories, introducing frequent camera motion, rapid zoom changes, and viewpoint shifts.
These differences are crucial for downstream policy learning, scene reconstruction, and motion prediction tasks.
Unlike prior datasets that isolate robotic or manual modalities, SurgTAP-3D explicitly distinguishes and includes both, enabling cross-domain learning and generalization analysis.
Specifically, we categorize collected videos according to their surgical modality to facilitate modality-specific training and benchmark tasks.

\subsubsection{Unified Action Categories} Previous datasets~\cite{nwoye2025cholectrack20} are often restricted to a single type of surgery, resulting in inconsistent or incompatible action labels across various datasets.
\change{To build a consistent taxonomy, we manually reviewed the action definitions and visual patterns in widely used surgical datasets, and defined five action categories that are common in practice and visually distinguishable: \textbf{dissection, tissue retraction, needle grasping, vessel clipping, and coagulation}.}  \change{For labeled source datasets, we map original labels to the closest category in our taxonomy. For unlabeled or incompatible videos, we manually assign categories based on visible instrument motion, tool--tissue interaction, and surgical context.}
\change{As summarized in Fig.~\ref{fig: data_statistics}, the unified action set includes dissection, tissue retraction, needle grasping, vessel clipping, and coagulation.}

\begin{table*}[t]
    \small
    \centering
    \caption{ \textbf{Description of SurgTAP-3D.} The dataset contains the video datasets from both robotic surgery and manual surgery.} 
    \setlength{\tabcolsep}{0.85mm}{
    \begin{tabular}{llcccccccc}
        \toprule
          \textbf{Type}&\textbf{Datasets}&\textbf{\hm{Surgery}} & \textbf{Clips} &  \textbf{Frames}  & \textbf{ Avg. Frames} \\
         \midrule
         \multirow{2}{*}{Robotic Surgery}
         &GraSP~\cite{ayobi2025pixel} & Prostatectomy &795&109939&138\\
         &SAR-RARP50~\cite{psychogyios2023sar} &Prostatectomy & 741 & 62822 & 85 \\
         \midrule
         \multirow{7}{*}{Manual Surgery}
         &SLAM~\cite{ye2025comprehensive}& Multiple procedures & 203 & 24609 & 121\\
         &SurgicalActions160~\cite{schoeffmann2018video} & Gynecology &26 &2194 & 84\\
         &AutoLaparo~\cite{wang2022autolaparo} & Hysterectomy &43 & 4684 & 109\\
         &heicoProctocolectomy~\cite{maier2021heidelberg} &Proctocolectomy&39&4538&116\\
         &heicoRectalResection~\cite{maier2021heidelberg} &Rectal resection&23&2300&100\\         
         &CholecT50~\cite{nwoye2022data} &Cholecystectomy & 2031 & 234894 & 116\\
         &MultiBypass140~\cite{lavanchy2024challenges} & Gastric bypass & 648 & 73662 & 114\\
         
         \bottomrule
    \end{tabular}
    }
        \label{tab: release_dataset}
        \vspace{-1.0em}
\end{table*}

Following these design principles, as presented in Tab.~\ref{tab: release_dataset}, SurgTAP-3D aggregates multiple high-quality datasets under a unified framework: 1) Robotic surgery: GraSP~\cite{ayobi2025pixel}, SAR-RARP50~\cite{psychogyios2023sar}, 2) Manual Surgery: SLAM~\cite{ye2025comprehensive}, SurgicalActions160~\cite{schoeffmann2018video}, AutoLaparo~\cite{wang2022autolaparo}, the proctocolectomy and rectal-resection subsets of HeiCo~\cite{maier2021heidelberg}, CholecT50~\cite{nwoye2022data}, and MultiBypass140~\cite{lavanchy2024challenges}. In total, we collect 4549 clips, 519,642 frames, and 1,376,987 trajectories.
\subsection{Dataset Preprocessing}\label{sec: data_preprocess}
First, we extract frames from the videos at their original frame rates. Based on the unified action categories, we select sequences that contain visually clear and well-defined actions. Since our motion tracking begins from prompts given in the first frame of each clip, we ensure that this initial frame clearly exposes the active instrument under the camera.
To guarantee the overall quality and reliability of the dataset, we apply a multi-stage filtering process prior to annotation.
We use a luminance-based filtering method to detect and remove frames with excessive brightness variation or strong overexposure. 
\change{This step removes only frames or short segments that are unreliable for pseudo-annotation, rather than normal illumination variations commonly observed in surgical videos.}
Next, we perform motion blur detection using gradient-based sharpness metrics, such as the Laplacian variance, to identify and exclude frames where tool or camera motion causes significant blurring. 
\change{After these filtering steps, the retained frames form temporally consistent sequences with stable lighting, clear visual features, and reliable image quality, providing a robust foundation for 3D reconstruction, trajectory extraction, and action annotation.}
 Finally, after filtering out noisy frames, we retain clips ranging from 5–30 seconds in duration, corresponding to 20–300 frames.
\subsection{Contact-aware Data Annotation Pipeline}\label{sec: pipeline}

\change{Given a sequence of $T$ images as input, our goal is to enable 3D contact-aware motion estimation for the video, generating labels including pointclouds, camera poses, intrinsics, 2D/3D tracking trajectories for the contact region. 
While our pipeline is generalizable to various scenes with arbitrary point prompts, we focus on the contact region, which is the local area where the active instrument (the main tool currently performing an action) directly interacts with surrounding tissues. 
Optionally, our pipeline supports known priors (depths, intrinsics, and poses) as input, together with images, making it more flexible and applicable for different dataset types.}

\begin{figure*}[t]
\centering
\includegraphics[width=0.98\linewidth]{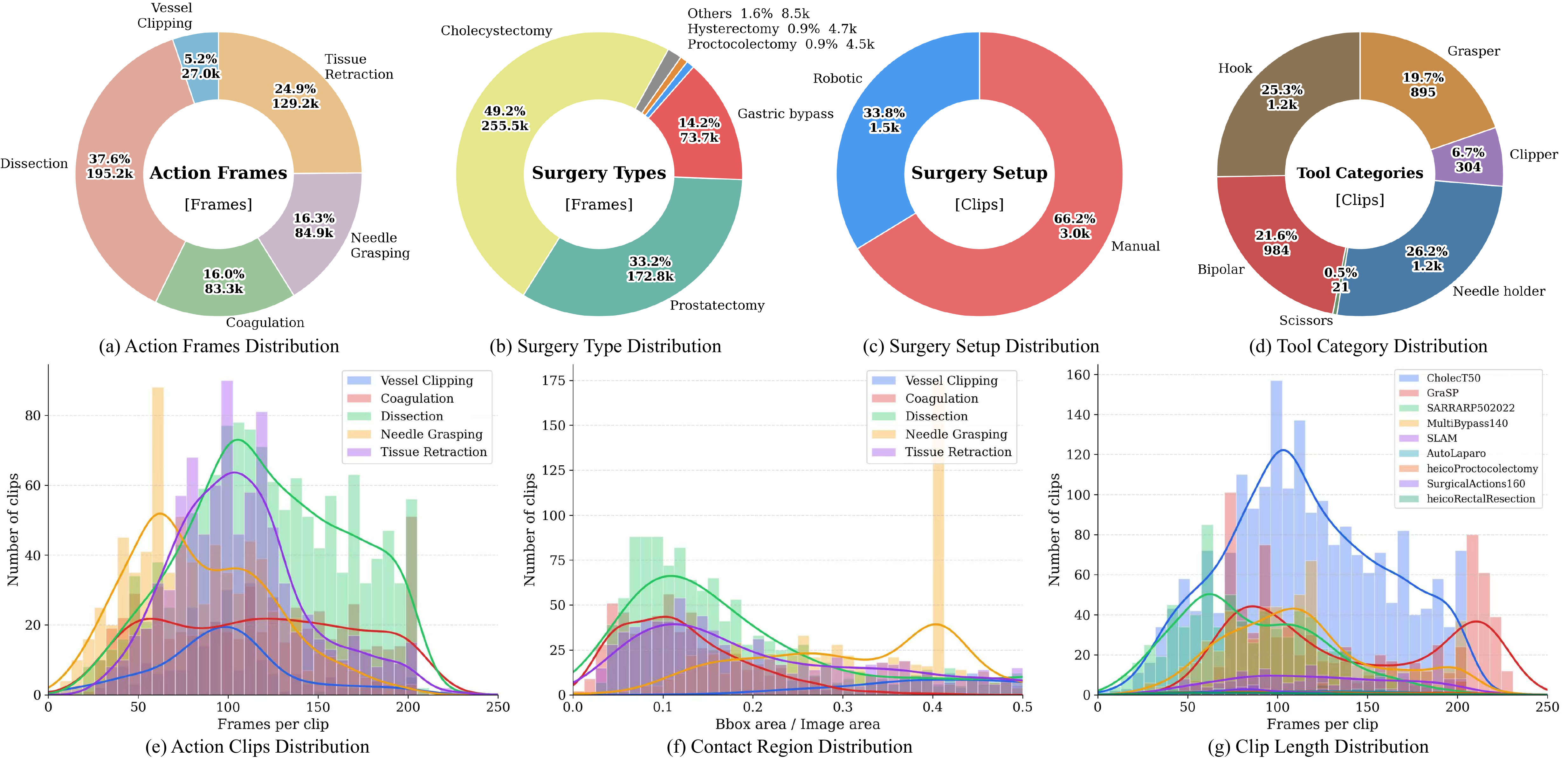}
\caption{\textbf{Data Statistics.} We analyze the data distribution of action frames, surgery types, surgical modality, and tool categories. We also visualize the clip lengths distribution in different actions, datasets, and the distribution of our labeled contact region.}
\vspace{-5pt}
\label{fig: data_statistics}
\end{figure*}
\subsubsection{Contact-aware Labeling and Segmentation} To efficiently annotate such regions, we design a simple yet effective one-frame labeling process. The annotator first draws a bounding box enclosing the contact region, including both the instrument tip and its immediate interaction area. Then, the annotator selects 3–5 positive points on the active instrument and 3–5 negative points on the background. These point prompts are provided to SAM2~\cite{ravi2024sam} to generate a precise segmentation mask of the active instrument for subsequent video tracking, generating active instrument masks for each frame as $\{\mathbf{M}_{t}^{seg} \}_{t=0}^{T-1}$.

\subsubsection{\change{Contact-region} 2D Tracking} Given $N$ query points $\bX_0^{2D}$ collected from the contact region, we utilize TAPNext~\cite{zholus2025tapnext} to obtain the 2D point tracking $\bX_t^{2D}$, visibility predictions $\bV_t$, confidence score $\bC_t$.
TAPNext~\cite{zholus2025tapnext} offers an online tracker to perform point tracking via a masked decoding manner. Unlike previous methods, which focus on correspondence matching and introduce complex inductive biases like cost volumes and iterative inference, TAPNext uses TRecViT as the backbone, which consists of ViT layers and state-space model layers for spatial and temporal processing, respectively. The 2D point tracking is formulated as:
\begin{equation}
    \change{\bX_t^{2D}, \bV_t, \bC_t = \text{TAPNext}(\bI_t, \bX_0^{2D}), \quad t=0, ..., T-1}.
\end{equation}

\subsubsection{Surgical 3D Reconstruction} 
Reconstructing the surgical scene is an inherently ill-posed and unconstrained problem, due to the challenges including dynamic instruments for intervention, deformable tissues, reflective surfaces, and illumination fluctuations. Traditional methods like ORB-SLAM, COLMAP will fail in such a highly dynamic scene, resulting in inaccurate camera poses and incomplete 3D geometry.
\change{For processing datasets, we adopt Endo3R~\cite{guo2025endo3r} to estimate depth maps $\bD_t^E$, camera poses $\bP_t$, and intrinsics $\bK$. 
Endo3R fits our setting because it is designed for endoscopic scenes, and our source videos are mostly uncalibrated monocular clips without stereo, kinematics, or ground-truth poses.
Unlike general depth models~\cite{piccinelli2025unidepthv2,wang2025moge} that predict per-frame depth only, Endo3R jointly estimates scale-consistent video depth and camera parameters, providing a stable geometry initialization for pseudo-3D annotation.}
It is proposed as a feed-forward 3D foundation model, with an uncertainty-aware dual-memory mechanism to enable long-term dynamic 3D reconstruction. To adapt to diverse surgery scenes, Endo3R designs a self-supervised training scheme for training across datasets even without ground truth. This process is mathematically formulated as:
\begin{equation}
    \bD_t^{E}, \bP_t, \bK = \text{Endo3R}(\bI_t), \quad t=0, ..., T-1.
\end{equation}
To further improve the depth fidelity, we adopt an off-the-shelf model~\cite{lin2025depth} to obtain depth $\bD_t^{M}$. 
We align the external depth predictions $\bD_t^{M}$ to the
original Endo3R depths $\bD_t^{E}$ by estimating a scale
$\alpha$ and shift $\beta$ through least-squares fitting over
valid pixels, yielding
$\Tilde{\bD}_t=\alpha\bD_t^{M}+\beta$.

\begin{figure*}[t]
\centering
\includegraphics[width=1.0\linewidth]{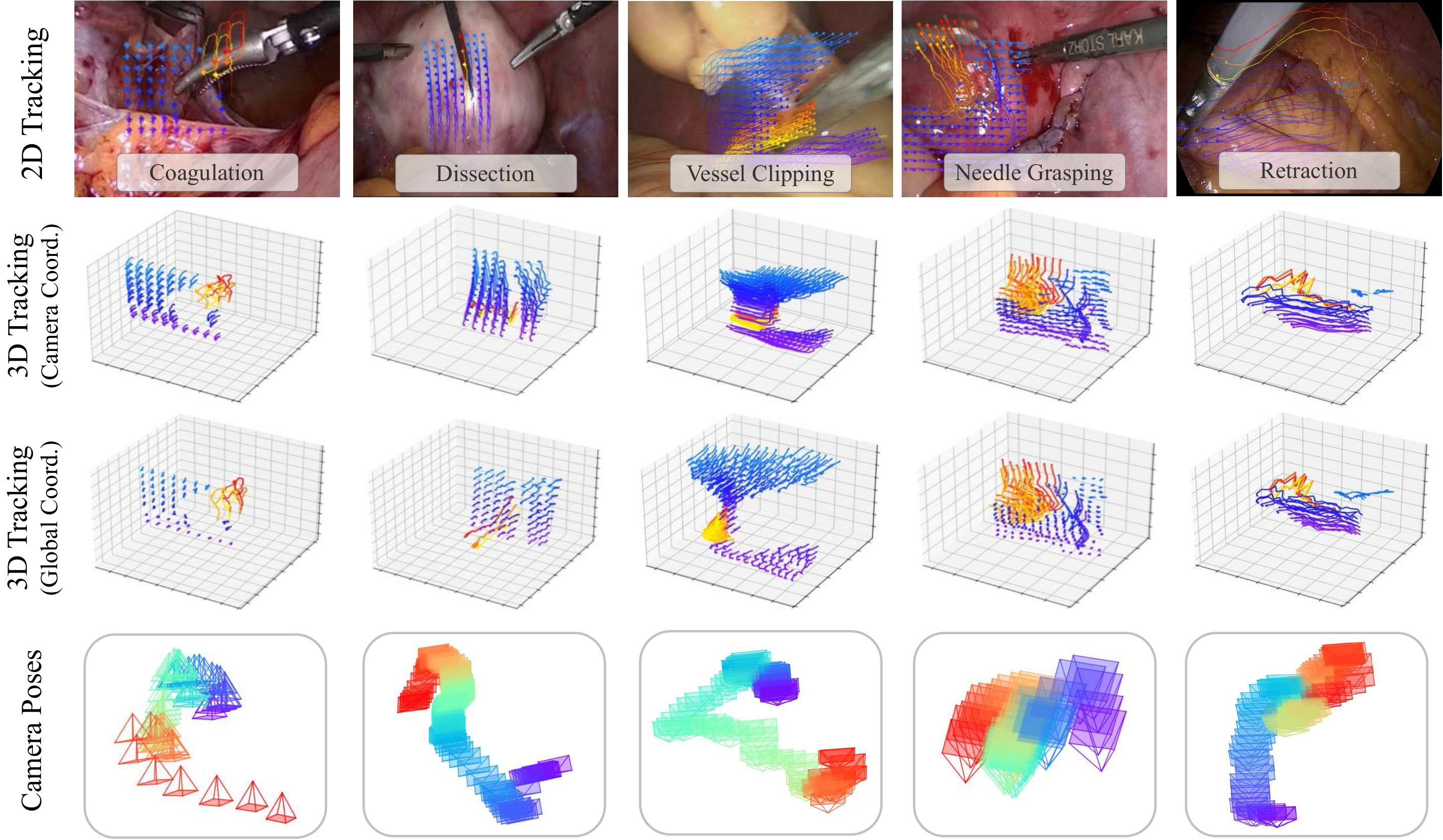}
\caption{\change{\textbf{Visualization of 2D and 3D Tracking in different actions.} We plot the point trajectories in 2D and 3D, with red-yellow trajectories for the active instrument and purple-blue trajectories for the tissues.}}
\label{fig: action_vis}
\end{figure*}
\subsubsection{\change{Dynamic Bundle Adjustment}}
\change{As a feed-forward method, Endo3R may still accumulate temporal drift over long clips. To mitigate long-term drift, we additionally employ a dynamic bundle adjustment inspired by~\cite{li2025megasam, teed2021droid}. Although DROID-SLAM is originally designed for online SLAM, we use its pretrained recurrent update module only as an optimization component over pre-collected clips. A dense bundle adjustment layer is constructed to jointly refine the keyframe camera poses, intrinsic parameters, and depths over a frame graph $G=(V, E)$, where $V$ denotes the set of keyframes and $E$ represents the edges between them, \change{where $(i, j)\in E$ denotes the connected edge between image $\bI_i$ and $\bI_j$.} The network then iteratively estimates the 2D correspondence fields $\hat{\bu}_{ij}$ and the confidence map $\bw_{ij}$, which are then used to refine keyframe camera poses, intrinsics, and depths.} 
Unlike previous methods~\cite{teed2021droid, li2025megasam} that use constant value to initialize, we initialize the poses, intrinsics, and depths with our predicted $\Tilde{\bD_t}, \bP_t, \bK$, which contain surgery-specified predictions and apply the \change{dynamic bundle adjustment} to refine these geometry parameters to achieve the drift correction for improved accuracy and robustness.

To optimize the intrinsics $\bK$, camera poses $\bP_t$, and video depth $\bD_t$, we minimize the reprojection cost between the predicted correspondence field and rigid motion:
\begin{equation}
     \mathcal{C}_{rigid} = \change{\sum_{(i,j) \in E}} \left\| \hat{\bu}_{ij} -  \pi(\bP_j^{-1}\bP_i \pi^{-1}(\bu_i, \bD_i, \bK^{-1}), \bK) \right\|_{\Sigma_{ij}}^2, 
\end{equation}
where $\Sigma_{ij} = \text{diag}(\bw_{ij} \odot \bn_{ij} \odot (1-\bM_{i}^{seg} (\bu_{i})))$ denotes the weights from the flow-induced confidence $\bw_{ij}$, estimated static mask $\bn_{ij}$ from network, and our annotated instrument masks $\bM_{i}^{seg}$. $\bu_i$ is the grid of pixel coordinates.
\change{During optimization, pixels with low correspondence confidence, strong non-rigid tissue motion, or annotated instrument regions receive lower weights in the reprojection residual. This down-weights unreliable dynamic regions and allows the bundle adjustment to rely more on geometrically stable regions for pose, intrinsic, and depth refinement.} The instrument masks also provide stable filtering from instruments, compensating for the uncertainty map when the instruments are in a temporary stop.
To further enhance the depth fidelity, we adopt a depth cost to constrain the optimized depth with the initialized depth $\Tilde{\bD}_t$.
\begin{equation}
    \mathcal{C}_{mono} = \sum_t \left\| \Tilde{\bD}_t - \bD_t \right\|^2_2.
\end{equation}
The aligned depth $\Tilde{\bD}_t$ remains fixed during
optimization, while $\bD_t$ denotes the optimized depth. The overall objective for \change{dynamic bundle adjustment} is therefore formulated as:
\begin{equation}
   \argmin_{\bK, \bP_t, \bD_t} \mathcal{C}_{rigid} + \mathcal{C}_{mono} .
\end{equation}
Moreover, when additional sensor priors (i.e., camera intrinsics, extrinsics, or depth measurements) are available, our framework can seamlessly incorporate them by introducing corresponding regularization terms. This design allows the pipeline to flexibly adapt to different data acquisition settings while maintaining geometric consistency and optimization stability.
Using the recovered geometry, we obtain the 3D tracking trajectory in the camera coordinate system $\bX_{t}^{3D, C}$ by lifting the 2D trajectory with the estimated depth $\bD_t$ and intrinsic matrix $\bK$. The 3D trajectory in \change{global coordinates} $\bX_{t}^{3D, W}$ is then computed by transforming $\bX_{t}^{3D,C}$ with the camera pose $\bP_t$:
\begin{align}
    \bX_{t}^{3D, C} &= \pi^{-1}(\bX_t^{2D}, \bD_{t} (\bX_t^{2D}) , \bK^{-1}), \\
    \bX_{t}^{3D, W} &= \bP_t\pi^{-1}(\bX_t^{2D}, \bD_{t} (\bX_t^{2D}) , \bK^{-1}).
\end{align}
\subsubsection{Global 4D Refinement}
After extracting the 3D geometry estimation, we perform a post-optimization step to enhance the robustness, accuracy, and temporal smoothness of the tracking results, aligning the consistency between point tracking and flow estimation. All trajectory-based losses are computed only over tracker-visible valid points.

\noindent\textbf{Flow Consistency Loss. }Specifically, based on the pose graph, we computed the optical flow $\bO_{ij}$ between the linked frame pairs using the off-the-shelf model~\cite{teed2020raft}. 
 To jointly refine both depth estimation and point tracking, we design two objectives, \changetwo{$\mathcal{L}_{depth}$ and $\mathcal{L}_{track}$ to optimize the depth consistency and 2D tracking consistency, respectively}.
 For each connected frame pair $(i,j)\in E$, we propagate the tracked
points in frame $i$ to frame $j$ using optical flow:
\begin{equation}
    \widetilde{\bX}_{ij}^{2D}
    =
    \bX_i^{2D}
    +
    \bO_{ij}(\bX_i^{2D}),
\end{equation}
where $\bO_{ij}$ denotes the optical flow from $\bI_i$ to $\bI_j$.
The 2D tracking consistency loss is then defined as:
\begin{equation}
    \mathcal{L}_{track}
    =
    \sum_{(i,j)\in E}
    \left\|
    \bX_j^{2D}
    -
    \widetilde{\bX}_{ij}^{2D}
    \right\|_2^2.
\end{equation}
To stabilize the reconstructed 3D trajectories, we further enforce
cross-frame depth consistency for the static region between the 3D trajectories reconstructed
from $i$ and the depth map of $j$ at the corresponding
flow-warped locations:
\begin{equation}
    \mathcal{L}_{depth}
    =
    \sum_{(i,j)\in E}
    \left\|
    \bn_{ij}\odot
    \left(
    d(\bP_j^{-1}\bX_i^{3D,W})
    -
    \bD_j(\widetilde{\bX}_{ij}^{2D})
    \right)
    \right\|_2^2,
\end{equation}
where $d(\cdot)$ extracts the depth (z-coordinate).
The overall flow consistency loss is:
\begin{equation}
    \mathcal{L}_{flow}
    =
    \mathcal{L}_{track}
    +
    \mathcal{L}_{depth}.
\end{equation}

\noindent\textbf{Surface Constraint Loss. }
To improve the robustness of 3D point tracking on surgical instruments, we leverage the inherent consistency of the instrument and design a surface constraint loss to optimize the 3D instrument points $\mathcal{I}$ lying directly to the 3D surface of the instrument.
This constraint encourages the optimized 3D keypoints to lie directly on the instrument surface, correcting drift that commonly arises in 2D tracking due to occlusion, specular highlights, or motion blur. 
For each frame, we first reconstruct the instrument surface pointcloud $\bS_{t}$ by backprojecting depth $\bD_t$ within the segmentation mask $\bM_{t}^{seg}$. We then enforce geometric consistency by minimizing the Chamfer-style distance between the predicted 3D keypoints and the instrument surface.
\begin{align}
\quad \bS_{t} &= \left\{
\pi^{-1}\!\left(\mathbf{u},\, \mathbf{D}_t(\mathbf{u}),\, \mathbf{K}^{-1}\right)
\;\middle|\;
\bM_{t}^{seg}(\mathbf{u}) = 1
\right\}, \\
    \mathcal{L}_{surf} &= \sum_t\sum_{n \in \mathcal{I}} \min_{\bp \in \bS_{t}} \left\| \bX^{3D,C}_{t, n} - \bp \right\|^2_2.
\end{align}

\begin{figure}[t]
\centering
\includegraphics[width=1.0\linewidth]{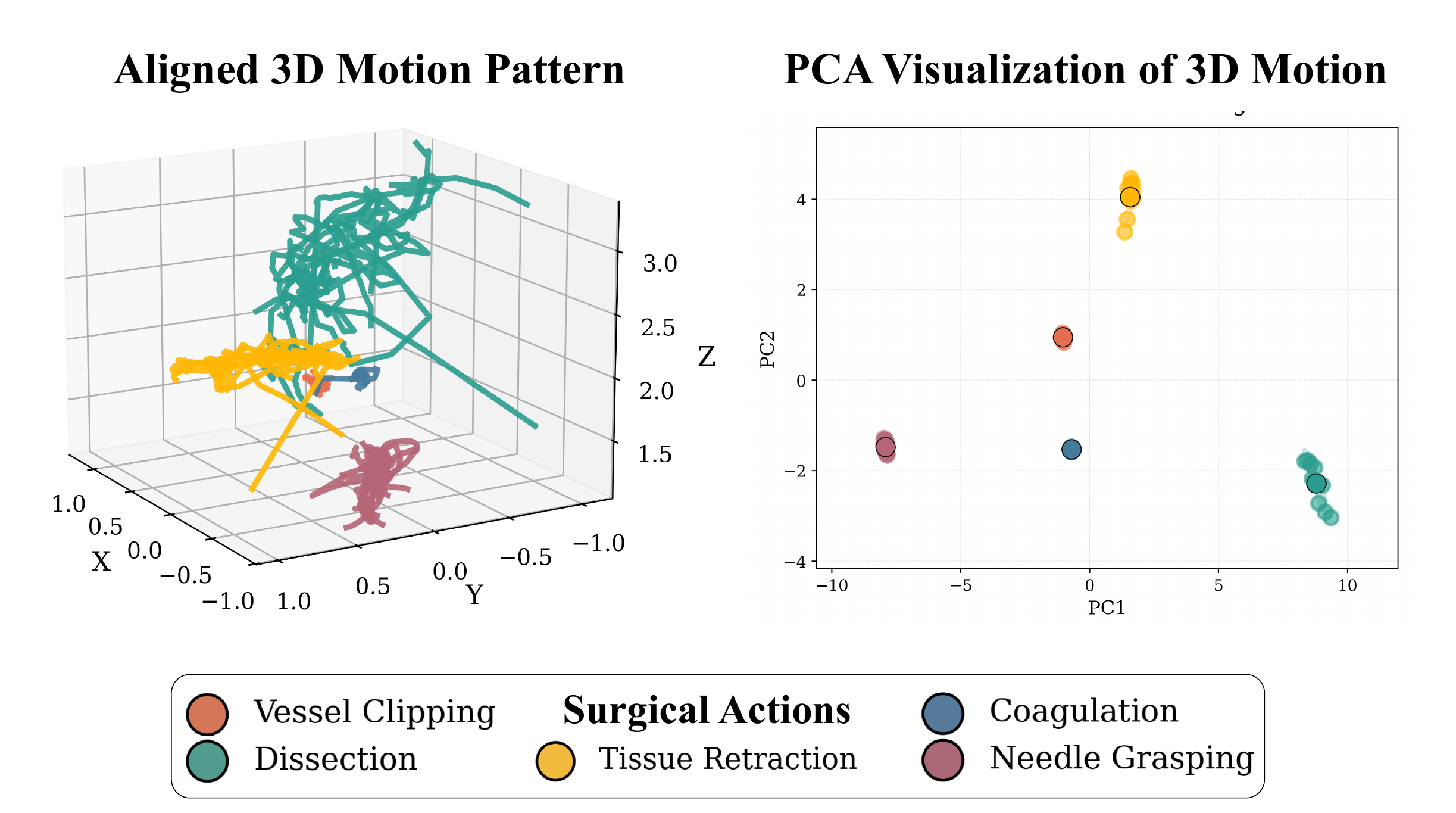}
\caption{\textbf{Correlation between Actions and 3D Trajectories.} We randomly sample ten clips per action and normalize their 3D instrument trajectories via spatial alignment. The projected mean trajectories form well-separated clusters, revealing a strong correlation between action categories and instrument motion patterns.}
\vspace{-10pt}
\label{fig: pca_vis}
\end{figure}

\begin{figure}[t]
\centering
\includegraphics[width=0.99\linewidth]{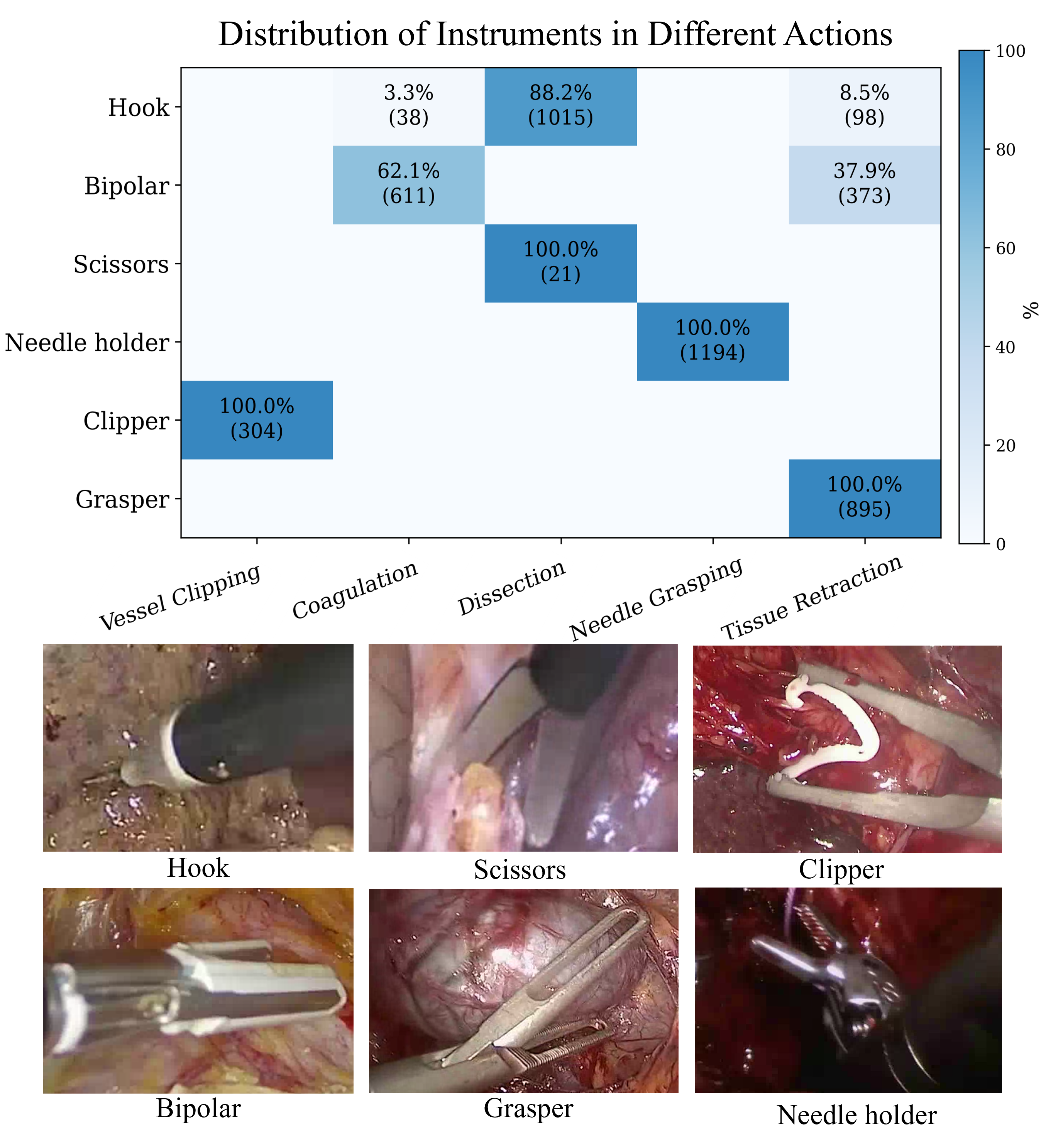}
\caption{\textbf{Correlation between Actions and Instrument Types.} We plot the heatmap to visualize the distribution of instrument types in different actions.}
\label{fig: action_ins}
\end{figure}
\noindent\textbf{\change{Articulated} ARAP Loss. }
To regularize the 3D trajectories of the instrument keypoints and ensure physically plausible motion, we introduce an \change{Articulated} As-Rigid-As-Possible (ARAP) constraint.
Unlike classical ARAP, which assumes global rigidity, surgical instruments contain rigid shaft and non-rigid articulating joints (e.g., wrist, grasper). Our formulation leverages multi-frame 3D keypoint trajectories to identify stable edges (pairs of keypoints) whose relative distances remain nearly constant across time. \changetwo{We define the reliability score as the inverse variance of the edge distances $r_{ij}$, and obtain the edges $\mathcal{E}$ with higher score than a threshold $\tau_q$, which correspond to rigid parts of the instrument (e.g., tool shaft):
\begin{align}
    r_{ij} &= \frac{1} {\text{Var}\left(\left\| \bX^{3D, C}_{t,i} - \bX^{3D, C}_{t,j} \right\|_2\right) + \epsilon}, \\  \mathcal{E} &= \{ (i, j) | r_{ij} > \tau_q \},
\end{align}where $\epsilon$ is a small constant for numerical stability.} For the rigid edges, we apply an ARAP-style penalty to suppress jitter in rigid regions and avoid over-constraining in articulated joints. 
\begin{equation}
    \mathcal{L}_{arap} = \sum_t \sum_{(i,j)\in\mathcal{E}} r_{ij} (\left\| \bX_{t,i}^{3D} - \bX_{t,j}^{3D} \right\|_2 - \left\| \bX_{0,i}^{3D} - \bX_{0,j}^{3D} \right\|_2)^2.
\end{equation}

\noindent\textbf{Temporal Smoothness Loss. }To improve the smoothness of 3D trajectories, we regularize their local temporal acceleration:
\begin{equation}
     \mathcal{L}_{dyn}
    =
    \sum_{\Delta\in W}
    \sum_{t=\Delta}^{T-1-\Delta}
    \left\|
    \bX^{3D,C}_{t+\Delta}
    -2\bX^{3D,C}_{t}
    +\bX^{3D,C}_{t-\Delta}
    \right\|_2^2,
\end{equation}
where $W$ denotes the temporal window used to compute local acceleration.

\noindent\changetwo{\textbf{2D Trajectory Regularization.}
Let $\widehat{\mathbf{X}}^{2D}_t$ denote the initial tracker
output and $\mathbf{X}^{2D}_t$ denote its optimized position.
We regularize the optimized trajectories with initial observations to prevent the optimized trajectories from drifting:
\begin{equation}
\mathcal{L}_{reg}
=
\sum_t
\left\|
\mathbf{X}^{2D}_{t}
-
\widehat{\mathbf{X}}^{2D}_{t}
\right\|_2^2.
\end{equation}}Finally, the overall optimization objective is formulated as:
\begin{equation}
\argmin_{\bD_t, \bX^{2D}_{t}} \lambda_1\mathcal{L}_{flow} + \lambda_2\mathcal{L}_{dyn} + \lambda_3\mathcal{L}_{surf} + \lambda_4\mathcal{L}_{arap} + \lambda_5 \mathcal{L}_{reg},   
\end{equation}
This is solved using the Adam optimizer with a learning rate of 0.05, with $\lambda_1=0.1,\lambda_2=0.1,\lambda_3=10,\lambda_4=100, \lambda_5=0.001$.
Through this optimization, the reconstructed 3D trajectories become spatially coherent, temporally smooth, and semantically consistent with the instrument regions.
\subsection{Data Analysis}\label{sec: data_stat}
\subsubsection{Statistics Overview} The SurgTAP-3D dataset contains 4,549 sequences and 519,642 frames, with each clip ranging from 5-30 seconds in duration, corresponding to 20-300 frames. 
As shown in Fig.~\ref{fig: data_statistics}, the dataset covers a wide variety of surgical scenarios. The top row of charts illustrates the distribution of action types, procedure types, surgical settings, and the instruments used, highlighting the diversity and balance of the collected data.
The second row provides further detail on the length of sequences across different actions and datasets. Additionally, the contact region analysis shows how much of the image is occupied by the main surgical activity, highlighting that the dataset focuses on the areas where tools and tissues interact most. This diversity and scale ensure that the dataset is a robust resource for developing 3D tracking methods that work across many different types of surgery.

\subsubsection{3D Tracking under Different Coordinates} As shown in Fig.~\ref{fig: action_vis}, we visualize the 3D tracking trajectories in both camera and \change{global coordinates}. While certain trajectories exhibit significant displacement in camera coordinates due to ego-motion, they exhibit smaller, more stable motion in \change{global coordinates}. This clearly demonstrates the critical importance of decoupling camera motion from scene motion to recover object trajectories, which is essential for downstream robotic tasks such as imitation learning, motion prediction, and manipulation planning.
\begin{figure*}[t]
\centering
\includegraphics[width=1.0\linewidth]{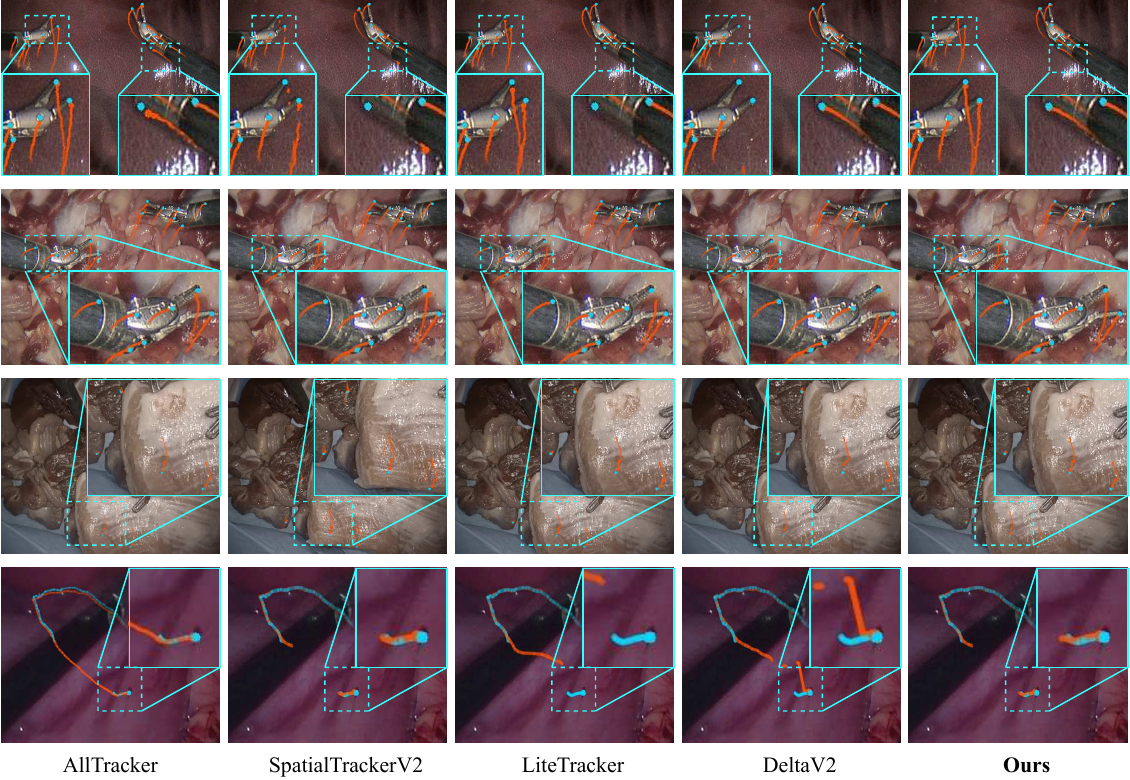}
\caption{\textbf{Qualitative Comparison} of 2D point tracking. We extensively compare with three types of point tracking methods, with \textcolor[rgb]{0.0, 0.705, 1.0}{blue points} as the ground truth trajectory \change{(from SurgPose~\cite{wu2025surgpose}, STIR~\cite{schmidt2025point}, and SurgT~\cite{cartucho2024surgt})} and \textcolor[rgb]{1.0, 0.295, 0.0}{orange points} as the predicted trajectory. The results exhibit our better robustness to occlusions, illumination variance, and instrument tracking.}
\label{fig: qual_track}
\end{figure*}

\subsubsection{Correlation between Actions and 3D Trajectories} To further understand the motion characteristics embodied in SurgTAP-3D, we analyze the correlation and distinction across action categories using aligned 3D tool trajectories. For each video, we normalize the raw trajectories to remove viewpoint and scale variations, and we aggregate them across all clips of the same action class. As presented in Fig.~\ref{fig: pca_vis}, by projecting the aligned mean trajectories into a low-dimensional motion embedding space using PCA, we observe that different actions form compact and well-separated clusters, indicating that each action exhibits a characteristic motion signature. This visualization not only reveals the inherent structure in surgical kinematics but also highlights the discriminative nature of the trajectories across cutting, grasping, pushing, and other manipulation types. Moreover, these motion priors provide a foundational resource for physically intelligent autonomous surgery. The clustered structure suggests that 3D trajectories can serve as compact, task-level representations for imitation learning, enabling policies to learn action-conditioned motion primitives with improved generalization.

\subsubsection{Correlation between Actions with Instruments} We additionally annotate instrument types for each video clip, covering six categories: hook, grasper, clipper, needle holder, scissors, and bipolar. As shown in Fig.~\ref{fig: action_ins}, the row-normalized distribution reveals a strong yet structured correlation between instrument types and surgical actions. 
Most instruments exhibit clear functional specialization: hooks and scissors are predominantly associated with dissection; needle holders are exclusively used for needle grasping; and graspers are primarily dedicated to tissue retraction. Clippers correspond closely to vessel clipping, whereas bipolars demonstrate broader usage across coagulation and tissue retraction, reflecting higher functional versatility.

This structured instrument–action coupling provides informative priors for high-level autonomous decision-making. First, it enables probabilistic modeling of tool–action compatibility, which can facilitate automatic tool selection and switch planning in robotic systems.
Second, the strong correlation between instrument identity and action category supports surgical state recognition by narrowing the hypothesis space of ongoing tasks based on observed tool usage. Finally, incorporating such instrument priors into decision-making pipelines can improve robustness in long-horizon autonomy by enforcing semantically consistent action transitions, thereby enhancing the reliability of policy learning and task-level planning in autonomous surgical systems.

\begin{table}[t]
\centering
\caption{\changetwo{\textbf{Annotation quality analysis across action classes.}
We report SR, IMC, TCR, and TSR in \% as indicators of
track survival, instrument-mask consistency, tracking confidence,
and temporal smoothness, respectively.}}
\label{tab:annotation_quality_action}
\setlength{\tabcolsep}{2.6mm}
\renewcommand{\arraystretch}{1.1}
\resizebox{\linewidth}{!}{
\begin{tabular}{lcccc}
\toprule
\textbf{Action Class}
& \textbf{SR}$\uparrow$
& \textbf{IMC}$\uparrow$
& \textbf{TCR}$\uparrow$
& \textbf{TSR}$\uparrow$\\
\midrule
Dissection
& 91.89 & 96.89 & 98.53 & 93.70  \\
Tissue Retraction
& 83.80 & 97.10 & 98.07 & 94.20  \\
Needle Grasping
& 91.46& 93.55 & 96.90 & 97.51  \\
Vessel Clipping
&88.21&   97.03&   99.45&  92.13 \\
Coagulation
& 86.71 & 95.35 & 98.72 & 92.88  \\
\midrule
\textbf{Average}
& 88.41 & 95.98 & 98.33 & 94.08 \\
\bottomrule
\end{tabular}
}
\vspace{-1mm}
\end{table}

\begin{figure}[t]
\centering
\includegraphics[width=1.0\linewidth]{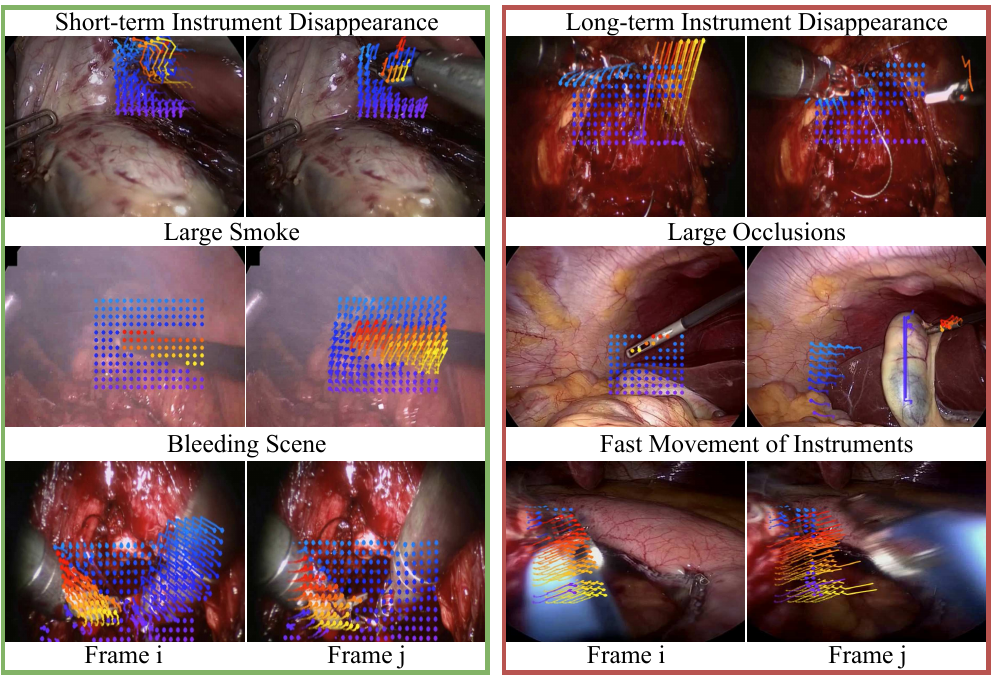}
\caption{\change{\textbf{Challenging-scene analysis of Annotation Quality.} \textcolor[rgb]{0.0, 0.55, 0.0}{Green box} shows robust cases, including short-term instrument disappearance ($<$10s), smoke, and bleeding. \textcolor[rgb]{0.75, 0.0, 0.0}{Red box} shows failure-prone cases, including long-term instrument disappearance ($>$10s), large occlusion, and fast instrument motion, where trajectories may drift or become unstable.}}
\label{fig: challenge-vis}
\end{figure}
\subsubsection{\change{Annotation Quality Assessment}}
We assess the generated pseudo-annotations using four consistency metrics. Survival Ratio (SR) measures the percentage of contact-initialized trajectories that remain valid. 
Instrument Mask Consistency (IMC) measures the percentage of instrument trajectories that remain inside the active-tool masks.
Temporal Smoothness Ratio (TSR) measures the percentage of valid temporal triplets whose second-order displacement is below a fixed threshold. Tracking Confidence Ratio (TCR) measures the percentage of trajectories with confidence scores above a fixed threshold. These metrics jointly assess trajectory validity, temporal stability, motion consistency, and semantic alignment with instrument regions.
\changetwo{Specifically, for trajectory $n$ at frame $t$, let $V_{t,n}\in\{0,1\}$
and $C_{t,n}$ denote its predicted visibility and tracking confidence,
and let $B_{t,n}\in\{0,1\}$ indicate whether $\bX_{t,n}^{2D}$ is finite
and inside the image. We denote $\mathcal{C}$ and $\mathcal{I}$ as the
trajectories initialized in the first-frame contact region and active-tool
mask, respectively. We further define the valid-observation indicator
$A_{t,n}=V_{t,n}B_{t,n}$, the valid-triplet indicator
$Q_{t,n}=A_{t-1,n}A_{t,n}A_{t+1,n}$, and the second-order displacement
$\Delta^2\bX_{t,n}^{2D}
=\bX_{t+1,n}^{2D}-2\bX_{t,n}^{2D}+\bX_{t-1,n}^{2D}$.
The four metrics are then defined as
\begin{align}
\mathrm{SR}
&=
\frac{\sum_{n\in\mathcal C}\sum_t A_{t,n}}
     {T|\mathcal C|},
\\
\mathrm{IMC}
&=
\frac{
\sum_{n\in\mathcal I}\sum_t
A_{t,n}\bM_t^{seg}(\bX_{t,n}^{2D})
}{
\sum_{n\in\mathcal I}\sum_t A_{t,n}
},
\\
\mathrm{TCR}
&=
\frac{
\sum_{n\in\mathcal C}\sum_t
A_{t,n}\mathbf{1}[C_{t,n}\ge\tau_c]
}{
\sum_{n\in\mathcal C}\sum_t A_{t,n}
},
\\
\mathrm{TSR}
&=
\frac{
\sum_{n\in\mathcal C}\sum_{t=1}^{T-2}
Q_{t,n}\mathbf{1}
[\|\Delta^2\bX_{t,n}^{2D}\|_2<\tau_s]
}{
\sum_{n\in\mathcal C}\sum_{t=1}^{T-2}Q_{t,n}
},
\end{align}
where $\tau_c=0.5$ and $\tau_s=5$ pixels. 
As shown in Tab.~\ref{tab:annotation_quality_action}, the pseudo-annotations exhibit high internal consistency across action classes.} 
Tissue retraction, vessel clipping, and coagulation show relatively lower SR due to stronger occlusion, deformation, abrupt tool motion, and cautery-induced appearance changes, which can temporarily invalidate contact-initialized tracks. 
IMC remains high, indicating good alignment between instrument trajectories and active-tool masks, while the slightly lower IMC in needle grasping is mainly caused by thin grasper tips and partial occlusion. 
The high TCR and stable TSR further indicate that retained trajectories are generally confident and temporally smooth after filtering. 
\changetwo{In the absence of ground truth, these metrics provide complementary indicators of annotation consistency and reliability rather than direct measures of geometric accuracy. 
They may also be affected by tracker-derived
visibility and confidence, segmentation quality, image boundaries, and
thresholds. 
Their consistently high values across
action classes support SurgTAP-3D as large-scale pseudo-annotations with
confidence indicators.}

\subsubsection{\change{Failure Analysis and Quality Control}}
\change{Although SurgTAP-3D achieves high overall annotation quality, it remains a pseudo-annotated dataset which may contain label noise under challenging visual conditions. As shown in Fig.~\ref{fig: challenge-vis}, our pipeline is generally robust to short-term instrument disappearance, smoke, and bleeding when sufficient visual correspondence remains available. However, failure-prone cases still exist, including long-term instrument disappearance, large occlusion, and fast instrument motion, where trajectories may disappear, drift, or become temporally unstable. These issues are more likely to occur in challenging interaction classes such as tissue retraction, needle grasping, and coagulation.}

\change{For quality control, we remove severely corrupted frames, initialize annotations with first-frame contact-region and active-tool prompts, and filter unreliable trajectories using visibility and confidence scores. The annotation cost is mainly from clip curation and first-frame prompt labeling, while dense 2D/3D trajectories are generated automatically by the pipeline. We also release confidence and visibility indicators to support filtering low-confidence trajectories for downstream usages.}

\section{Experiments}
\begin{table*}[htbp]
    \centering
    \caption{\textbf{Comprehensive 3D and 2D surgical motion tracking Benchmark.} We extensively evaluate the 2D and 3D point tracking for both instrument (SurgPose~\cite{wu2025surgpose}) and tissue surface (SurgT~\cite{cartucho2024surgt} and STIR~\cite{schmidt2025point}). 
    }
    \label{tab:tracking_main}
    \resizebox{0.95\textwidth}{!}{
    \setlength{\tabcolsep}{2.0pt}
    \renewcommand{\arraystretch}{1.1}
    \begin{tabular}{llcccccccccccccc}
    \toprule
    
    \multicolumn{14}{c}{\textit{2D Tracking}} \\
    \midrule
    \multirow{2}{*}{\textbf{Type}} & \multirow{2}{*}{\textbf{Method}} 
    & \multicolumn{3}{c}{\textbf{SurgT}~\cite{cartucho2024surgt}} 
    & \multicolumn{3}{c}{\textbf{SurgPose}~\cite{wu2025surgpose}} 
    & \multicolumn{3}{c}{\textbf{STIR}~\cite{schmidt2025point}}  
    & \multicolumn{3}{c}{\textbf{Average}} \\
    \cmidrule(lr{0.1em}){3-5} \cmidrule(lr{0.1em}){6-8} \cmidrule(lr{0.1em}){9-11} \cmidrule(lr{0.1em}){12-14}
    & & AJ$_{2D}$$\uparrow$ & APD$_{2D}$$\uparrow$ & OA$\uparrow$ 
    & AJ$_{2D}$$\uparrow$ & APD$_{2D}$$\uparrow$ & OA$\uparrow$ 
    & AJ$_{2D}$$\uparrow$ & APD$_{2D}$$\uparrow$ & OA$\uparrow$ 
    & AJ$_{2D}$$\uparrow$ & APD$_{2D}$$\uparrow$ & OA$\uparrow$\\
    \midrule
\multirow{5}{*}{{I}}
& BootsTAPIR~\cite{doersch2024bootstap}   & 67.85    & 73.42    & \textbf{99.10}    & 47.28    & 57.79    & 91.65    & 12.33    & 16.47    & 96.74   & 42.49    & 49.23    & 95.83    \\
    & CoTrackerv3~\cite{karaev2025cotracker3}         & 73.32    & 79.02    & 98.75    & 49.40    & 61.84    & 83.89    & 19.34    & 24.23    & 96.51 & 47.35    & 55.03    & 93.05    \\
    & AllTracker~\cite{harley2025alltracker}          & 68.11    & 75.30    & 96.59     & 53.59    & 64.19    & \textbf{97.93}    & 17.87    & 22.57    & \textbf{99.48}   & 46.52    & 54.02    & \textbf{98.00}    \\
    & TAPNext~\cite{zholus2025tapnext}          & 74.98    & \underline{80.52}    & 98.18    & \underline{60.35}    & \underline{69.78}    & 96.40    & 18.89    & 23.46    & 95.46 & \underline{51.41}    & \underline{57.92}    & 96.68    \\
    & LiteTracker~\cite{karaoglu2025litetracker} & 71.20&76.71 &98.61 & 51.63 & 62.68 & 86.20 &\textbf{21.37}&\textbf{26.76}&95.30 & 48.07 & 55.38 & 93.37 \\
    
    \midrule
    \multirow{4}{*}{{II}}
    & DELTA~\cite{ngo2024delta}      & \underline{75.25}     & 80.07   & 97.23    & 51.71   & 62.61  & 80.29   & 18.71    & 24.20    & 95.71  & 48.56 & 55.63 & 91.08    \\
    & DELTAv2~\cite{ngo2025deltav2}         & 73.75 & 79.21    & 96.03 & 51.20 & 61.97 & 77.43  &  16.92 & 21.77 & 90.62 & 47.29 & 54.32 & 88.03    \\
    & SpatialTracker~\cite{xiao2024spatialtracker} & 70.95    & 75.96    & 98.20    & 49.16    & 60.86    & 89.79    & 18.85   & \underline{25.01}   & 91.01   & 46.32 & 53.94 & 93.00 \\
    & \change{Online3D~\cite{hayoz2024}} & \change{44.96} & \change{50.38} & \change{95.30} & \change{4.74} & \change{8.32} & \change{96.59} & \change{14.71} & \change{19.58} & \change{93.64} & \change{21.47} & \change{26.09} & \change{95.18}\\
    \midrule
    \multirow{3}{*}{{III}}
    & TAPIP3D~\cite{zhang2025tapip3d}                        & 74.83    & 79.82   & 98.37    & 38.06    & 50.11    & 95.03    & 17.68    & 24.26    & 94.72 & 43.52    & 51.40    & 96.04    \\
    & SpatialTrackerv2~\cite{xiao2025spatialtrackerv2}            & 66.66   & 71.62    & \underline{98.89}    & 42.58    & 61.41    & 81.13    & 18.36   & 23.23    & 95.80  & 42.53 & 52.09 & 91.94    \\
    & \textbf{Ours}  & \textbf{77.68}& \textbf{81.43} &98.63 & \textbf{60.78} &\textbf{70.12} & \underline{96.77} & \underline{20.93} &24.82 &\underline{97.03} & \textbf{53.13} & \textbf{58.79} & \underline{97.48}\\
    \midrule
    \multicolumn{14}{c}{\textit{3D Tracking}} \\
    \midrule
    \multirow{2}{*}{Type} & \multirow{2}{*}{Method} 
    & \multicolumn{3}{c}{\textbf{SurgT}~\cite{cartucho2024surgt}} 
    & \multicolumn{3}{c}{\textbf{SurgPose}~\cite{wu2025surgpose}} 
    & \multicolumn{3}{c}{\textbf{STIR}~\cite{schmidt2025point}}
    & \multicolumn{3}{c}{\textbf{Average}} \\
    \cmidrule(lr{0.1em}){3-5} \cmidrule(lr{0.1em}){6-8} \cmidrule(lr{0.1em}){9-11} \cmidrule(lr{0.1em}){12-14}
    & & AJ$_{3D}$$\uparrow$ & APD$_{3D}$$\uparrow$ & OA$\uparrow$ 
    & AJ$_{3D}$$\uparrow$ & APD$_{3D}$$\uparrow$ & OA$\uparrow$ 
    & AJ$_{3D}$$\uparrow$ & APD$_{3D}$$\uparrow$ & OA$\uparrow$
    & AJ$_{3D}$$\uparrow$ & APD$_{3D}$$\uparrow$ & OA$\uparrow$\\
    \midrule
\multirow{5}{*}{{I}}
& BootsTAPIR~\cite{doersch2024bootstap}             & 87.81    & 91.77    & 99.08    & 30.28    & 40.47    & 91.30    & 27.18    & 34.60    & \underline{98.29}& 48.42    & 55.61    & 96.22    \\
    & CoTrackerv3~\cite{karaev2025cotracker3}         & 90.17    & 93.99    & 98.84    & 35.51    & 47.26    & 84.22    & 28.16    & 34.76    & 98.21   & 51.28    & 58.67    & 93.76    \\
    & AllTracker~\cite{harley2025alltracker}          & 88.30    & 94.13    & 96.38    & 33.26    & 43.80    & \textbf{97.49}    & 30.19    & 36.87    & \textbf{99.57}   & 50.58    & 58.27    & \textbf{97.81}    \\
    & TAPNext~\cite{zholus2025tapnext}          & \underline{91.53}    & \underline{94.61}    & \textbf{99.26}    & \underline{42.64}    & \underline{53.43}    & 96.08    & 32.89    & 40.59    & 96.95  & \underline{55.69}    & \underline{62.88}    & 97.43    \\
    & LiteTracker~\cite{karaoglu2025litetracker} & 87.62    & 91.30    & 98.54    & 36.11    & 48.40    & 84.35    & 30.76    & 38.46    & 92.23  & 51.50    & 59.39    & 91.71    \\
    \midrule
    \multirow{4}{*}{{II}}
    & DELTA~\cite{ngo2024delta}   & 88.00  &  91.11 &   97.31      & 39.66 & 49.18   & 86.49   & 32.52    & 41.95    & 92.75 & 53.39    & 60.75    & 92.18   \\
    & DELTAv2~\cite{ngo2025deltav2}         & 85.20 & 89.14    & 95.90 &38.03 &48.53 &80.81 &   31.14& 39.05 & 87.97 & 51.45 & 58.91 & 88.23\\
    & SpatialTracker~\cite{xiao2024spatialtracker}   & 76.30    & 82.51    & 98.22    & 36.83    & 45.21    & 80.85    & 25.35    & 33.30    & 95.50   & 46.16    & 53.67    & 91.52    \\
    &  \change{Online3D~\cite{hayoz2024}} & \change{73.93} & \change{80.27} & \change{95.34} & \change{12.42} & \change{15.80} & \change{98.53} & \change{30.88} & \change{40.06} & \change{95.38} & \change{39.07} & \change{45.38} & \change{96.41} \\
    \midrule
    \multirow{3}{*}{{III}}
    & TAPIP3D~\cite{zhang2025tapip3d}    & 86.46    & 89.91    & 98.36   & 37.29   & 48.19   & 90.95    & 29.63    & 38.75    & 92.03   & 51.12   & 58.95    & 93.78   \\
    & SpatialTrackerv2~\cite{xiao2025spatialtrackerv2}       & 88.03    & 91.93    & 99.21    & 30.88    & 41.97    & 85.76    & \underline{33.45}    & \underline{42.55}    & 92.94    & 50.78    & 58.82    & 92.64    \\
    & \textbf{Ours} &  \textbf{91.78} & \textbf{94.82} & \underline{99.23} &  \textbf{44.16} & \textbf{56.71}& \underline{96.13} &  \textbf{33.79} & \textbf{42.77} &97.05&\textbf{56.57} & \textbf{64.77} & \underline{97.47} \\ 
    \bottomrule
    \end{tabular}
    }
\end{table*}

To comprehensively evaluate the proposed pipeline, we establish a benchmark that systematically compares our method with state-of-the-art approaches for both 2D and 3D point tracking on public datasets covering soft tissue and instrument keypoint tracking. 
In addition, we assess the dynamic reconstruction capability by benchmarking camera pose estimation and video depth estimation against existing methods. 

\subsection{2D and 3D Point Tracking}

\subsubsection{Baselines}
To evaluate the tracking performance on surgery datasets, we build a large benchmark to include the current state-of-the-art methods for 2D and 3D tracking. We classify the current methods into three types, type I refers to the 2D tracking methods taking RGB as input (BootsTAPIR~\cite{doersch2024bootstap}, CoTracker~\cite{karaev2024cotracker}, AllTracker~\cite{harley2025alltracker},
TAPNext~\cite{zholus2025tapnext},
LiteTracker~\cite{karaoglu2025litetracker}), type II refers to the methods taking RGBD data as input for tracking under camera coordinate, using the depths from monocular depth model (DELTA~\cite{ngo2024delta}, DELTAv2~\cite{ngo2025deltav2}, SpatialTracker~\cite{xiao2024spatialtracker}) \change{or depths from stereo depth model (Online3D~\cite{hayoz2024})}; type III includes the methods taking RGBD and camera poses as input to achieve 3D tracking in \change{global coordinate} (TAPIP3D~\cite{zhang2025tapip3d}, SpatialTrackerv2~\cite{xiao2025spatialtrackerv2}).

We extensively evaluate the point tracking results on both 2D and 3D tracking. 
\change{Following previous work~\cite{cartucho2024surgt,schmidt2025point}, the 3D tracking evaluation uses stereo-derived reference trajectories, while our tracking pipeline remains monocular at inference time. Although these 3D trajectories are stereo-derived evaluation instead of perfect physical ground truth, we assume that the accuracy of stereo labels and stereo calibrations is sufficient for 3D tracking validation. Therefore, the limitation exists in the accuracy of stereo calibrations and manual labeling.} For evaluation, we compute the 3D trajectory $\bX^{3D}$ by first computing the disparity from the 2D tracking results from left and right images, then projecting the point into 3D space using stereo calibration parameters $\bK_L$
and stereo baseline $b$:
\begin{equation}
    \bz = f_x b / (\bu_L-\bu_R),
    \quad
    \bX^{3D} = \pi^{-1}(\bX^{2D}_{left}, \bz, \bK_L^{-1})
    .
\end{equation}

\subsubsection{Benchmark Datasets}
We adopt three public datasets to validate both instrument and tissue tracking: SurgPose~\cite{wu2025surgpose}, SurgT~\cite{cartucho2024surgt}, and STIR~\cite{schmidt2025point}. All three datasets contain stereo images and 2D tracking ground truth, facilitating the evaluation for both 2D and 3D tracking. For each dataset, we conduct preprocessing to rectify the stereo images and use the stereo calibration parameters.
Here we give a brief introduction to the benchmark datasets. SurgPose~\cite{wu2025surgpose} is a dataset for instance-level 3D surgical instrument pose estimation and tracking, containing approximately 120,000 annotated instrument instances across 6 tool categories, collected from surgical instrument motions over ex-vivo tissue using dVRK. 
Each instance is annotated with semantic keypoints defining functional structures such as shafts, wrists, jaws, and tips. Ground-truth 3D keypoints are derived via high-precision stereo triangulation from synchronized stereo endoscope streams.

SurgT~\cite{cartucho2024surgt} is a standardized benchmark for soft-tissue tracking in robotic surgery. The dataset comprises 157 stereo endoscopic videos from 20 clinical cases, totaling over 1 hour of footage, with bounding-box annotations for regions of interest on deforming tissue. It includes stereo camera calibration parameters, enabling 3D reconstruction and evaluation of both 2D and 3D tracking metrics. 

The STIR Challenge focuses on point tracking for deformable tissue in surgery. This dataset features in-vivo and ex-vivo da Vinci robotic surgery videos marked with indocyanine green tattoos visible only in near-infrared. It includes hundreds of sequences with 2D/3D point trajectories, with key challenges under deformations, tool occlusions, and camera motion. \change{We follow the standard evaluation protocol in STIR paper, using the full sequence without truncation.}

\subsubsection{Metrics}
We adopt the metrics for 3D tracking following TAPvid-3D~\cite{koppula2024tapvid}, measuring the quality of the predicted 3D point trajectories (APD$_{3D}$), the ability to predict point visibility (OA), and 3D Average Jaccard (AJ$_{3D}$). For 2D tracking evaluation, we follow TAP-Vid~\cite{doersch2022tap} to adopt APD$_{2D}$, AJ$_{2D}$ and OA for evaluating the 2D tracking. 

\begin{table}[tbp]
\scriptsize
\centering
\caption{\textbf{Pose Comparison on SCARED~\cite{allan2021stereo}.}}
\label{tab: pose_results}
\resizebox{\linewidth}{!}{
\setlength{\tabcolsep}{2.5mm}
 \begin{tabular}{lccc}
        \toprule
         \multirow{1}{*}{\textbf{Method}} &
         \textbf{ATE} $\downarrow$ & \textbf{RPE$_r$} $ \downarrow$& \textbf{RPE$_t$} $\downarrow$  \\
         \midrule
         Endo-SfM~\cite{ozyoruk2021endoslam} &0.157&0.252&0.259\\
         AF-SfM~\cite{shao2022self}&0.125&0.235&0.241 \\
         EndoDAC~\cite{cui2024endodac} &0.124&0.223&0.233 \\
         Robust~\cite{hayoz2023learning} &0.131&0.241&0.245\\
         Endo3R~\cite{guo2025endo3r} &0.112 & 0.201 & 0.228\\
         \change{SpatialTrackerV2~\cite{xiao2025spatialtrackerv2}} & \change{0.110}&\change{0.217}&\change{0.239}\\
         \textbf{Ours} & \textbf{0.103} & \textbf{0.195} & \textbf{0.217}\\
        \bottomrule
    \end{tabular}
}
\end{table}

\begin{table*}[tbp]
\scriptsize
\centering
\caption{\textbf{Quantitative comparison with state-of-the-art video depth estimation methods}.}
\label{tab: depthresults}
\resizebox{\linewidth}{!}{
\setlength{\tabcolsep}{1.3mm}
\begin{tabular}{@{}lcccccccccc@{}}
\toprule
\multirow{2}{*}{\textbf{Method}} &
\multicolumn{5}{c}{\textbf{SCARED}~\cite{allan2021stereo}} &
\multicolumn{5}{c}{\textbf{Hamlyn}~\cite{mountney2010three,stoyanov2010real,pratt2010dynamic}} \\
\cmidrule(l{2pt}r{2.5pt}){2-6} \cmidrule(l{2pt}r{2.5pt}){7-11}
& Abs Rel $\downarrow$ & Sq Rel $\downarrow$ & RMSE $\downarrow$ & RMSE log $\downarrow$& $\delta < 1.25$ $\uparrow$ & Abs Rel $\downarrow$ & Sq Rel $\downarrow$ & RMSE $\downarrow$ & RMSE log $\downarrow$& $\delta < 1.25$ $\uparrow$ \\
\midrule
Monodepth2~\cite{godard2019digging}      & 0.432  &3.548&4.704&0.431& 0.425 & 0.379 & 9.318&20.472&0.403&0.439 \\
Endo-SfM~\cite{ozyoruk2021endoslam}      & 0.241 & 0.865&2.286&0.267& 0.585 & 0.252 & 4.335&14.430&0.268&0.628 \\
AF-SfM~\cite{shao2022self}               & 0.257& 0.960&2.162&0.291&0.573&0.286& 5.715&15.895&0.301&0.508\\
EndoDAC~\cite{cui2024endodac}            &  0.242& 0.934&2.014&0.275&0.584&0.275& 5.557&15.669&0.288&0.519 \\
Transfer~\cite{budd2024transferring}        & 0.297& 1.207&2.561&0.319&0.561 & 0.281& 5.790&15.936&0.312&0.504 \\
DA-V2~\cite{yang2025depth}               & 0.313& 1.425&2.839&0.453&0.508 & 0.334& 7.713&19.548&0.362&0.461 \\
VDA~\cite{chen2025video}                 & 0.291& 1.186&2.447&0.296&0.647 & 0.315& 7.492&19.231&0.347&0.476 \\
Endo DM~\cite{recasens2021endo}          & 0.203& 0.651&2.063&0.245&0.612 & 0.216& 4.639&14.799&0.273&0.619 \\
Monst3R~\cite{zhang2024monst3r}          & 0.198& 0.539&1.965&0.234&0.626 & 0.198& 4.193&15.221&0.241&0.645 \\
Endo3R~\cite{guo2025endo3r}                           & 0.124 & 0.227 & 1.209 & 0.135 &  0.839 & 0.170 &3.139 &11.569&0.196&0.707  \\
\change{SpatialTrackerV2~\cite{xiao2025spatialtrackerv2}} & \change{0.142} & \change{0.291} & \change{1.360} & \change{0.152} & \change{0.791} & \change{0.161} & \change{3.428} & \change{14.120} & \change{0.201} & \change{0.674} \\

\change{UniDepthV2~\cite{piccinelli2025unidepthv2}} & \change{0.203} & \change{0.617} & \change{1.863} & \change{0.210} & \change{0.654} & \change{0.277} & \change{4.290} & \change{15.542} & \change{0.294} & \change{0.565} \\

\change{MoGeV2~\cite{wang2025moge}} & \change{0.196} & \change{0.688} & \change{1.869} & \change{0.196} & \change{0.704} & \change{0.236} & \change{3.901} & \change{14.652} & \change{0.278} & \change{0.643} \\

\change{DepthAnythingV3~\cite{lin2025depth}} & \change{0.131} & \change{0.248} & \change{1.294} & \change{0.143} & \change{0.838} & \change{0.176} & \change{3.597} & \change{14.979} & \change{0.202} & \change{0.652} \\
\textbf{Ours}                            & \textbf{0.103} & \textbf{0.210} & \textbf{1.127} & \textbf{0.124} & \textbf{0.853} & \textbf{0.156} & \textbf{2.986} & \textbf{10.713} & \textbf{0.184} & \textbf{0.732}\\
\bottomrule
\end{tabular}
}
\end{table*}

\subsubsection{Results and Analysis}
As reported in Tab.~\ref{tab:tracking_main}, we conduct extensive comparisons with current state-of-the-art methods from two perspectives: 2D tracking and 3D tracking. The results demonstrate that our proposed pipeline significantly outperforms the three representative categories of existing approaches.
As illustrated in Fig.~\ref{fig: qual_track}, we visualize representative 2D point tracking results, where blue points denote the ground-truth trajectories and orange points indicate the predicted trajectories. Point tracking in surgical scenes is particularly challenging due to intrinsic characteristics of the environment, including rapid instrument motion, homogeneous surface textures, and severe occlusions caused by tissue manipulation.

Under these conditions, tracking methods such as LiteTracker~\cite{karaoglu2025litetracker} and DeltaV2~\cite{ngo2025deltav2} frequently lose track when large occlusions occur (e.g., row 4 in Fig.~\ref{fig: qual_track}). Methods that rely on reconstruction-based priors, such as SpatialTrackerV2, exhibit degraded performance in instrument tracking near object boundaries, where fast motion and motion blur lead to premature track termination (rows 1 and 2 in Fig.~\ref{fig: qual_track}).
\change{As a stereo-based reconstruction pipeline, Online3D~\cite{hayoz2024} performs much worse on SurgPose~\cite{wu2025surgpose}, indicating that reconstructing and tracking fast-moving instruments remains difficult for 4D Gaussian Splatting-based methods.}
In contrast, our method achieves robust and accurate tracking across all challenging scenarios, maintaining tight alignment with ground truth without trajectory loss or noticeable drift.

\subsection{Video Depth Estimation}
In addition to the tracking performance, video depth estimation is also crucial for obtaining accurate 3D tracking results. 
\subsubsection{Baselines}
We compare with both monocular depth estimation methods under surgical scene (Monodepth2~\cite{godard2019digging}, Endo-SfM~\cite{ozyoruk2021endoslam}, Af-SfMLearner~\cite{shao2022self}, EndoDAC~\cite{cui2024endodac}, EndoDM~\cite{recasens2021endo}) and large 3d foundation models in general computer vision (DepthAnythingv2~\cite{yang2025depth}, Video Depth Anything~\cite{chen2025video}, and Monst3R~\cite{zhang2024monst3r})
\subsubsection{Datasets}
We validate the depth estimation on two commonly used datasets with ground truth: SCARED~\cite{allan2021stereo} and Hamlyn~\cite{mountney2010three,stoyanov2010real,pratt2010dynamic}. The Stereo Correspondence and Reconstruction of Endoscopic Data (SCARED) dataset provides high-resolution stereo endoscopic videos acquired from ex-vivo porcine abdominal anatomy. It includes calibrated stereo pairs, ground-truth depth maps derived from structured light, and accurate camera poses, forming a standard benchmark for evaluating depth estimation and 3D reconstruction methods in minimally invasive surgery.

The Hamlyn~\cite{mountney2010three,stoyanov2010real,pratt2010dynamic} dataset contains in-vivo surgical sequences captured during minimally invasive procedures. Depth ground truth is obtained via multi-view reconstruction with calibrated setups. Hamlyn exhibits challenging surgical scenes, with appearance variations, dynamic lighting, occlusions, and tissue deformation, providing a complementary evaluation of robustness and generalization in clinical scenarios.
\subsubsection{Metrics}
We follow~\cite{guo2025endo3r} to evaluate the monocular depth estimation performance by using five commonly used metrics: Abs Rel, Sq Rel, RMSE, RMSE log, $\delta<$1.25.

\subsubsection{Results and Analysis}
As reported in Tab.~\ref{tab: depthresults}, \change{we compare our method with both surgical depth methods and recent general depth/3D foundation models, including SpatialTrackerV2~\cite{xiao2025spatialtrackerv2}, UniDepthV2~\cite{piccinelli2025unidepthv2}, MoGeV2~\cite{wang2025moge}, and DepthAnythingV3~\cite{lin2025depth}. 
Endo3R~\cite{guo2025endo3r} consistently outperforms these baselines on most metrics, supporting its use as the geometry backbone for surgical videos. Built on this initialization, our depth refinement improves local geometry, while dynamic bundle adjustment jointly optimizes depth, camera poses, and intrinsics by suppressing unreliable dynamic regions. As a result, our full pipeline achieves the best overall performance, reducing Abs Rel from 0.124 to 0.103 on SCARED and from 0.170 to 0.156 on Hamlyn compared with Endo3R.}

\begin{table*}[t]
\centering
\vspace{-1mm}

\begin{minipage}[t]{0.49\linewidth}
\centering
\captionof{table}{\change{\textbf{Component ablation} of the annotation pipeline. Starting from Endo3R, we progressively add modules to evaluate their contributions to depth and pose estimation.}}
\label{tab:abl-module-r2}
\setlength{\tabcolsep}{0.9mm}
\renewcommand{\arraystretch}{1.1}
\resizebox{\linewidth}{!}{
\begin{tabular}{lccccc}
\toprule
\textbf{Setting}
& \textbf{Abs Rel$\downarrow$}
& \textbf{$\delta<1.25$$\uparrow$}
& \textbf{ATE$\downarrow$}
& \textbf{RPE$_r$$\downarrow$}
& \textbf{RPE$_t$$\downarrow$} \\
\midrule
Endo3R only
& 0.124 & 0.839 & 0.112 & 0.201 & 0.228 \\
+ Depth refine
& 0.115 & 0.847 & 0.112 & 0.201 & 0.228 \\
+ Dynamic BA
& 0.108 & 0.850 & 0.103 & 0.195 & 0.217 \\
+ Global 4D refine
& 0.103& 0.853 &  0.103 & 0.195 & 0.217 \\
\bottomrule
\end{tabular}
}
\end{minipage}
\hfill
\begin{minipage}[t]{0.49\linewidth}
\centering
\captionof{table}{\change{\textbf{Loss-term ablation} in global 4D refinement by removing each loss term from the full objective.}}
\label{tab:abl-loss-r2}
\setlength{\tabcolsep}{0.35mm}
\renewcommand{\arraystretch}{1.1}
\resizebox{\linewidth}{!}{
\begin{tabular}{lccccccc}
\toprule
\textbf{Setting}
& \textbf{AJ$_{2D}\uparrow$}
& \textbf{APD$_{2D}\uparrow$}
& \textbf{AJ$_{3D}\uparrow$}
& \textbf{APD$_{3D}\uparrow$}
& \textbf{OA$\uparrow$}
& \textbf{Abs Rel$\downarrow$}
& \textbf{$\delta<1.25$$\uparrow$} \\
\midrule
Full
& 53.13 & 58.79  & 56.57 & 64.77 & 97.47 &  0.103& 0.853  \\
    w/o $\mathcal{L}_{flow}$
    & 51.87 & 58.32 & 56.08 & 63.37 & 97.46 & 0.108 & 0.850 \\

    w/o $\mathcal{L}_{dyn}$
    & 52.89 & 58.62 & 56.36 & 64.23 & 97.47 & 0.103 & 0.853 \\

    w/o $\mathcal{L}_{surf}$
    & 52.48 & 58.35 & 56.40 & 64.07 & 97.44 & 0.103 & 0.853 \\

    w/o $\mathcal{L}_{arap}$
    & 52.54 & 58.40 & 56.22 & 63.68 & 97.46 & 0.103 & 0.853 \\

    w/o $\mathcal{L}_{reg}$
    & 51.57 & 57.83 & 55.41 & 62.80 & 97.28 & 0.105 & 0.853 \\
    \bottomrule
\end{tabular}
}

\end{minipage}
\vspace{-1mm}
\end{table*}

\subsection{Camera Pose Estimation}
\subsubsection{Datasets}
We follow previous work~\cite{guo2025endo3r} to validate the pose estimation results on the SCARED dataset, which has ground truth camera poses and intrinsics.
\subsubsection{Metrics}
To validate the pose accuracy, we follow previous methods~\cite{guo2025endo3r} to evaluate with metrics including Absolute Trajectory Error (ATE) and Relative Pose Error (RPE).
\subsubsection{Results and Analysis}
We report the pose accuracy comparison result in Tab.~\ref{tab: pose_results}. It shows that our method outperforms the previous approaches in all the metrics. Given the coarse results from Endo3R, our method further conducts \change{dynamic} bundle adjustment to refine the estimated intrinsics, extrinsics, and depths. Therefore, the pose accuracy is improved over the feed-forward methods, with noises to be effectively mitigated.

\subsection{Ablation Study}
\subsubsection{Ablation Study on Components of the Annotation Pipeline}
\change{We conduct the ablation to analyze the contribution of each component in the pipeline.
As shown in Tab.~\ref{tab:abl-module-r2}, Endo3R initialization provides a strong geometric prior and outperforms the recent general depth models (see Tab.~\ref{tab: depthresults}). 
Building on this initialization, depth refinement improves local geometric accuracy. Dynamic BA further enhances both depth and pose estimation by suppressing unreliable dynamic regions, leading to more stable joint optimization of scene geometry and camera motion. Finally, global 4D refinement achieves the best overall performance by enforcing long-range spatiotemporal consistency between trajectories and geometry.}

\begin{figure*}[t]
\centering
\includegraphics[width=1.0\linewidth]{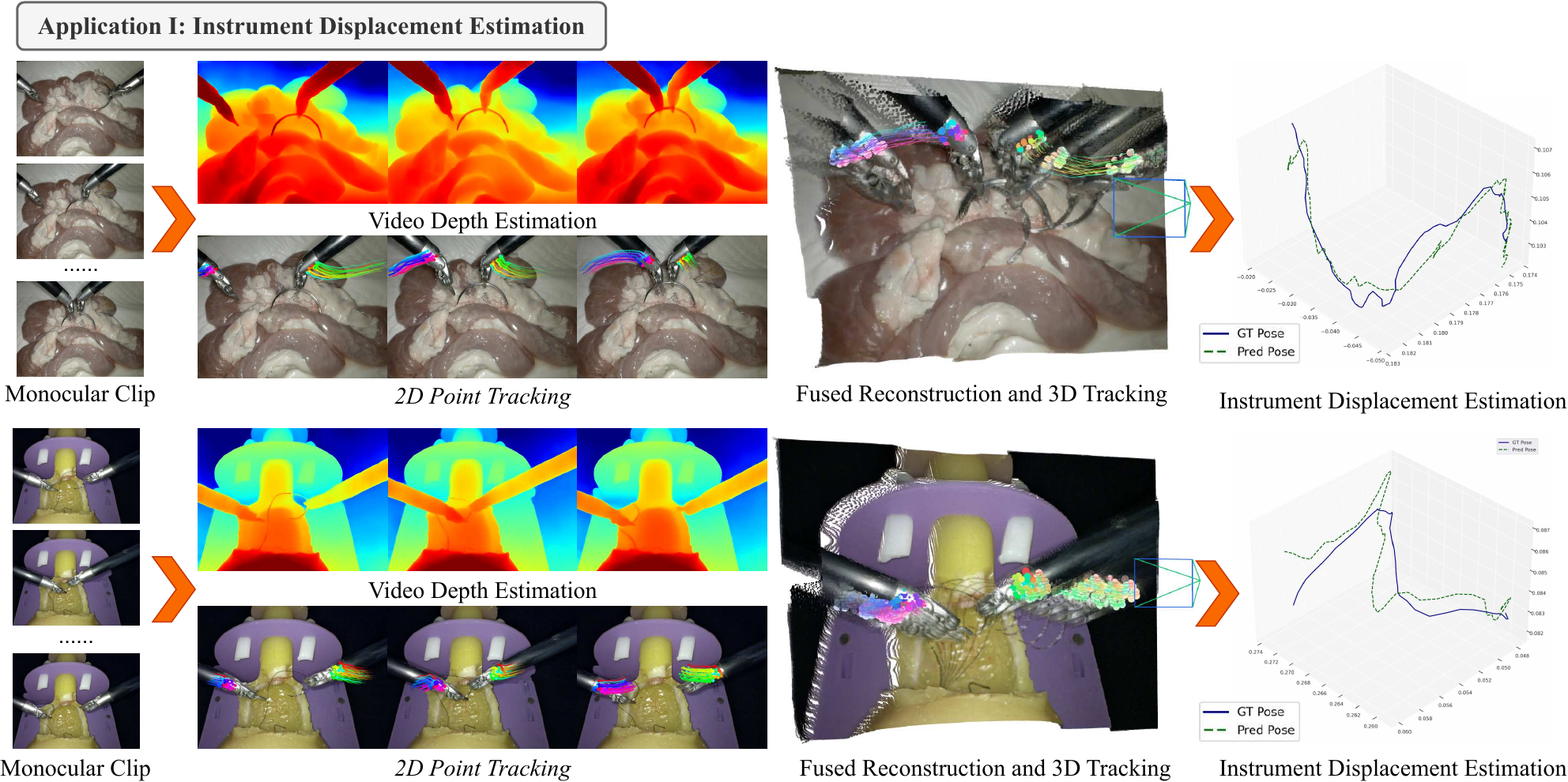}
\caption{\textbf{Real-world Application for Instrument Displacement Estimation.} Given the predicted 3D trajectory, the instrument displacement is estimated by computing the centroids of the instrument trajectories at each timestep. To validate the effectiveness, we collect \change{50 sequences} from dVRK with paired video clips and kinematic data. It shows that our framework enables accurate and robust instrument displacement estimation. }
\label{fig: app_1}
\end{figure*}

\subsubsection{Ablation Study on Losses in Global 4D Refinement}
\change{We conduct the ablation to analyze the contribution of each loss term in the global 4D refinement. As shown in Tab.~\ref{tab:abl-loss-r2}, the full objective achieves the best overall performance on both 2D and 3D tracking metrics. Removing $\mathcal{L}_{reg}$ causes a large degradation, showing the importance of preventing the optimized trajectories from drifting away from image evidence. 
Removing $\mathcal{L}_{flow}$ also leads to clear performance drops and worsens depth metrics, showing that flow consistency helps maintain cross-frame correspondence and geometry consistency. 
The temporal smoothness loss $\mathcal{L}_{dyn}$ provides smaller but consistent gains, mainly stabilizing trajectory optimization.
The surface constraint $\mathcal{L}_{surf}$ and articulated ARAP regularization $\mathcal{L}_{arap}$ further improve 3D tracking accuracy, validating the benefit of constraining instrument trajectories with surface and local-rigidity priors. Overall, these results show that the full combination of losses provides the most stable objective for pseudo-3D trajectory refinement.}

\section{Applications}
In this section, we present several application tasks to validate the utility and effectiveness of the SurgTAP-3D dataset. 
First, we validate our pipeline in real-world dVRK surgical settings by estimating instrument tip displacement, which is a critical task for autonomous manipulation. 
Second, we leverage our recovered geometry to facilitate deformable surface reconstruction and novel view synthesis. By providing reliable spatial priors, our method eliminates the dependency on traditional SfM techniques like COLMAP for 3DGS, while simultaneously improving reconstruction fidelity in dynamic surgical scenes. Finally, we demonstrate the value of SurgTAP-3D as a large-scale training dataset by reporting significant performance gains in state-of-the-art tracking models after fine-tuning on our data.

\subsection{Instrument Displacement Estimation}

Precise estimation of instrument displacement is a fundamental requirement for surgical automation and high-fidelity motion analysis. By leveraging pixel-wise 3D tracking across multiple points on the instrument tip, our framework inherently facilitates the recovery of tool translation and movement trajectories over time.
Specifically, we model the 3D trajectories as a set of points rigidly attached to the surgical instrument. To estimate per-frame displacement, we compute the centroid shift between consecutive frames using the top trajectories with the highest confidence scores at each time step. This strategy suppresses noisy correspondences while preserving rigid motion consistency. The resulting formulation converts raw 3D point-tracking outputs into pseudo-scale displacement estimates, yielding a stable and interpretable representation of the instrument’s motion path within the surgical workspace.

\begin{figure*}[t]
\centering
\includegraphics[width=1.0\linewidth]{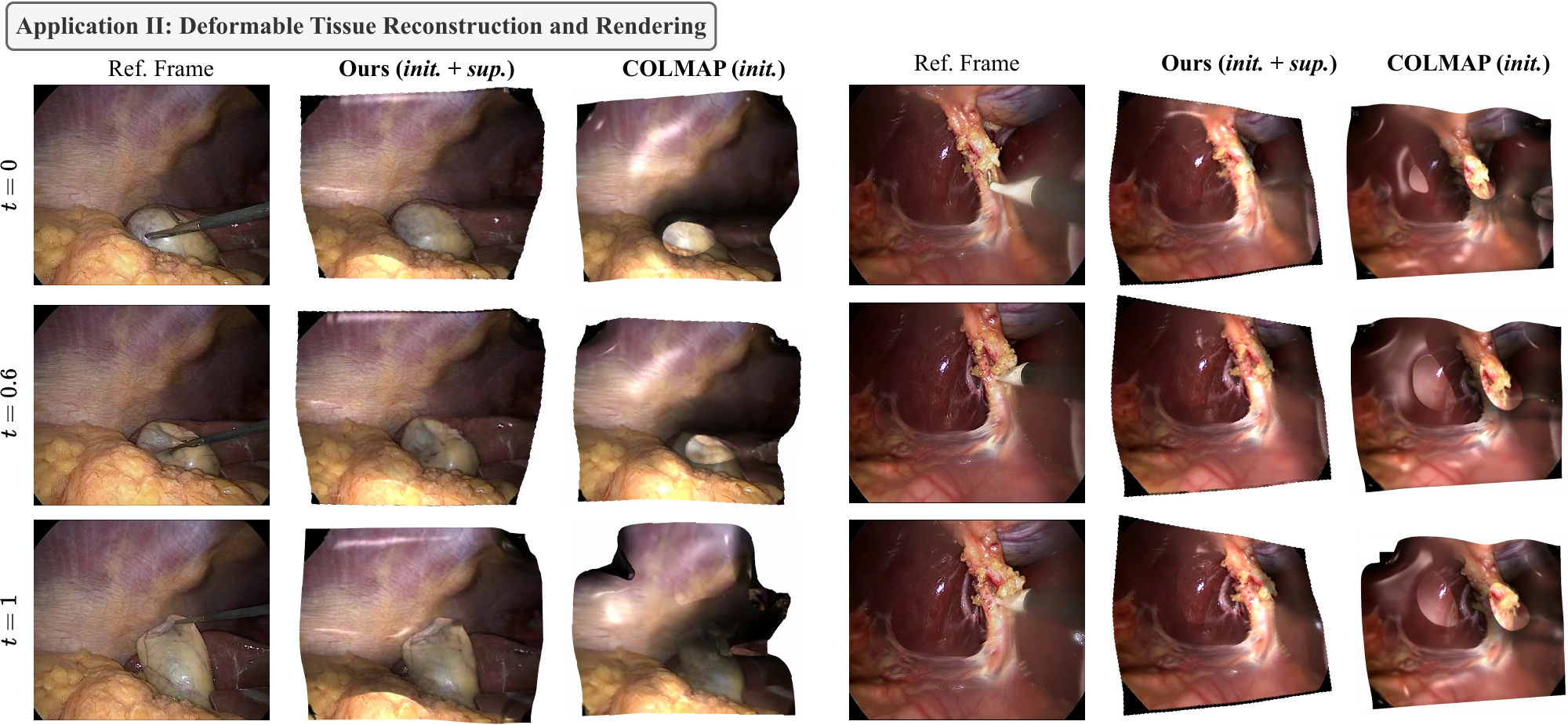}
\caption{\textbf{Real-world Application for Deformable Tissue Reconstruction and Rendering.} Given the reconstructed results, we train the 4D Gaussian Splatting model for deformable surface reconstruction and novel view synthesis for high-fidelity rendering. Compared with 4DGS with vanilla COLMAP initialization, our reconstructed point clouds, poses, and intrinsics yield significant improvement in reconstruction results.}
\label{fig: app_2}
\end{figure*}

\changetwo{Specifically, we backproject the visible 2D trajectories
using the associated depth and camera intrinsics and define the predicted
instrument position $\widehat{\mathbf{p}}_t$ as the centroid of the top K valid
3D points. Because $\widehat{\mathbf{p}}_t$ is recovered up to an unknown scale, we independently align each estimated displacement trajectory to its kinematic reference by estimating:
\begin{equation}
(s^{*},\mathbf{R}^{*},\mathbf{t}^{*})
=
\arg\min_{s,\mathbf{R},\mathbf{t}}
\sum_{t\in\mathcal{T}}
\left\|
\mathbf{p}_t-
\left(
s\mathbf{R}\widehat{\mathbf{p}}_t+\mathbf{t}
\right)
\right\|_2^2,
\end{equation}
where $\mathcal{T}$ denotes the valid synchronized frame pairs.
The Sim(3) parameters are obtained using the closed-form Umeyama solution.
For comparison, we validate all baselines using the same input sequences, Sim(3) alignment, and evaluation protocol by computing ATE and translational RPE$_t$ in meters from the aligned
trajectories. No camera-to-robot extrinsic calibration is available or used.}

\noindent \textbf{Real-surgery Data Collection.}
\change{We collect 50 paired monocular video-kinematic sequences from a dVRK robotic platform. The video stream is used as the input to our pipeline, while synchronized dVRK kinematics provide reference measurements for evaluation.} These sequences present significant challenges, including frequent occlusions, specular highlights, and motion blur. For each sequence, we tracked multiple 3D points on the instrument tip and reconstructed the displacement sequence by calculating the frame-to-frame positional changes of the tracked cluster. This setup allows for a quantitative assessment of both the absolute tracking accuracy and the temporal consistency of the recovered motion under realistic surgical conditions.

\begin{table}[tbp]
\scriptsize
\centering
\caption{\change{\textbf{Quantitative results for instrument displacement estimation} on the 50 sequences of self-collected paired video--kinematic dVRK dataset (unit: m).}}
\label{tab:instrument}
\resizebox{\linewidth}{!}{
\setlength{\tabcolsep}{4.8mm}
\begin{tabular}{lcc}
\toprule
\textbf{Method}
& \textbf{ATE}$\downarrow$
& \textbf{RPE}$_t\downarrow$ \\
\midrule
LiteTracker~\cite{karaoglu2025litetracker}
&0.0035 & 0.0015  \\
DELTAv2~\cite{ngo2025deltav2}
& 0.0040 & 0.0022 \\
SpatialTrackerV2~\cite{xiao2025spatialtrackerv2}
& 0.0064 & 0.0008 \\
\textbf{Ours}
& \textbf{0.0022}
& \textbf{0.0005} \\
\bottomrule
\end{tabular}
}
\end{table}

\begin{table}[tbp]
\scriptsize
\centering
\caption{\textbf{Quantitative Comparison} for Novel View Synthesis on SurgTAP-3D dataset. ``Init." refers to initialization. ``Sup." refers to depth supervision.}
\label{tab:nvs_compare}
\resizebox{\linewidth}{!}{
\setlength{\tabcolsep}{3.5mm}
\begin{tabular}{cccc}
\toprule
 \textbf{Init. + Sup.} &\textbf{PSNR}$\uparrow$ &
\textbf{SSIM}$\uparrow$ & \textbf{LPIPS}$\downarrow$  
\\
\midrule
 COLMAP + DAv2  & 27.82 & 0.8410 & 0.2161\\
 COLMAP + Ours &  27.51 &  0.8359 & 0.2196 \\
 Ours & \textbf{28.16}  &\textbf{0.8548} &\textbf{0.1310}\\
\bottomrule
\end{tabular}
}
\end{table}

\noindent \textbf{Results and Analysis.}
As illustrated in Fig.~\ref{fig: app_1}, our pipeline effectively processes monocular sequences to extract high-fidelity 3D trajectories. The visualized results confirm that our framework maintains spatial and temporal consistency, even during rapid or jerky instrument maneuvers.
\change{To quantitatively evaluate the displacement accuracy, we compare with different types of methods for 3D tracking (LiteTracker~\cite{karaoglu2025litetracker}, DELTAv2~\cite{ngo2025deltav2}, and SpatialTrackerv2~\cite{xiao2025spatialtrackerv2}) and report the Absolute Trajectory Error (ATE) and the translational Relative Pose Error (RPE$_t$) in Tab.~\ref{tab:instrument}.
The results show that our method consistently outperforms the previous approaches, demonstrating that our framework provides more accurate and temporally consistent instrument displacement estimation from uncalibrated monocular videos. 
The low RPE$_t$ indicates accurate local displacement estimation with reduced frame-to-frame drift, which is important for virtual fixture generation and surgical skill assessment~\cite{guo2024mt}.}

\subsection{Deformable Surface Reconstruction and Rendering}
The high-fidelity camera poses, scale-consistent depth maps, and temporally coherent 3D point trajectories provided by our SurgTAP-3D dataset enable deformable surface reconstruction and high-quality novel view rendering. Unlike prior surgical datasets that lack reliable temporal geometric priors, SurgTAP-3D facilitates the evaluation of 4D reconstruction methods in real-world, in-vivo surgical environments.

\noindent\textbf{Results and Analysis.}
To validate the effectiveness of our geometry estimation, we select 10 representative scenes from SurgTAP-3D for downstream application evaluation. We adopt Endo-4DGS~\cite{huang2024endo} for deformable surface reconstruction. Endo-4DGS represents the scene using Gaussian Splatting and models temporal deformation via a multi-layer perceptron (MLP), enabling efficient optimization and real-time novel view synthesis.

Most existing 3DGS methods rely on geometric priors (camera intrinsics, poses, and sparse point clouds), which are typically obtained through COLMAP~\cite{schonberger2016structure}. However, in surgical scenarios, COLMAP frequently fails due to textureless tissue surfaces, significant illumination variations, specular reflections, and the absence of rigid structural constraints~\cite{guo2024free}.

In contrast, our framework is designed to handle the highly dynamic surgical scene to provide dense geometry estimations. To validate the effectiveness of our method to this application, we replace COLMAP by directly providing accurate, scale-consistent camera poses and dense geometric priors tailored to dynamic surgical scenes. This results in more stable initialization and improved reconstruction quality for 3DGS-based methods. As reported in Tab.~\ref{tab:nvs_compare}, we compare our initialization with the default COLMAP-based pipeline. The results demonstrate that our geometry estimation substantially improves reconstruction fidelity and novel view rendering performance, particularly in challenging dynamic regions. As shown in Fig.~\ref{fig: app_2}, the qualitative result further demonstrates our consistent improvement over the vanilla 3DGS setting, exhibiting the effectiveness of our method for the downstream deformable reconstruction application.

\subsection{Finetuning for Models}
Given our datasets with large-scale and high-fidelity surgery action sequences, we obtain high-fidelity geometry reconstructions and motion tracking results. To validate the effectiveness of the SurgTAP-3D dataset, we additionally conduct a validation task to finetune CoTrackerv3~\cite{karaev2025cotracker3} and DELTAv2~\cite{ngo2025deltav2} on the train split of our dataset.

\noindent \textbf{Results and Analysis. }As illustrated in Table \ref{tab:finetune}, fine-tuning on our dataset yields consistent performance improvements for point tracking across both the SurgT \cite{cartucho2024surgt} and SurgPose \cite{wu2025surgpose} benchmarks. We attribute this enhancement to the comprehensive nature of SurgTAP-3D, which captures synchronized tracking data for both instruments and tissues across a diverse range of surgical environments. These results demonstrate the superior generalization capabilities and transferability of our dataset as a foundational resource for surgical motion analysis.

\begin{table}[tbp]
\scriptsize
\centering
\caption{\textbf{Quantitative Results} for Finetuning CoTrackerV3~\cite{karaev2025cotracker3} and DELTAv2~\cite{ngo2025deltav2} on our dataset. }
\label{tab:finetune}
\resizebox{\linewidth}{!}{
\setlength{\tabcolsep}{1.0mm}
\begin{tabular}{ccccc}
\toprule
\multirow{2}{*}{\textbf{Settings}} 
& \multicolumn{2}{c}{\textbf{SurgT}~\cite{cartucho2024surgt}} 
& \multicolumn{2}{c}{\textbf{SurgPose}~\cite{wu2025surgpose}} \\
\cmidrule(lr){2-3} \cmidrule(lr){4-5}
& \textbf{AJ}$_{2D}\uparrow$ & \textbf{APD}$_{2D}\uparrow$ 
& \textbf{AJ}$_{2D}\uparrow$ & \textbf{APD}$_{2D}\uparrow$ \\
\midrule
CoTrackerv3~\cite{karaev2025cotracker3}     & 73.32 & 79.02& 49.40 & 61.84 \\
CoTrackerv3~\cite{karaev2025cotracker3} w/ ft. & \textbf{74.14} & \textbf{80.25} &  \textbf{50.99} & \textbf{62.53} \\
\midrule
DELTAv2~\cite{ngo2025deltav2} & 73.75 & 79.21 & 51.20 & 61.97\\
DELTAv2~\cite{ngo2025deltav2} w/ ft. & \textbf{74.57} & \textbf{80.11} & \textbf{52.75} & \textbf{62.86} \\ 

\bottomrule
\end{tabular}
}
\end{table}

\section{Discussion}

\change{SurgTAP-3D addresses the lack of large-scale, geometrically consistent 3D surgical data for learning-based surgical perception. It converts uncalibrated monocular surgical videos into pseudo-annotated 4D trajectories without requiring dedicated robotic hardware or calibrated capture setups.}

\change{Our pipeline combines feed-forward reconstruction, depth, tracking, and segmentation priors with geometric refinement. Foundation priors provide scalability, while dynamic bundle adjustment and global 4D optimization improve camera/depth consistency, instrument rigidity, articulated motion coherence, and temporal smoothness. The resulting trajectories are spatially consistent in both camera and \change{global coordinate} systems, supporting relative 3D motion analysis.
SurgTAP-3D expands existing surgical tracking benchmarks and addresses the need for large-scale, cross-surgery, and contact-aware 3D datasets. It covers in-vivo procedures across multiple surgery types and action categories, and provides contact-aware dense 2D/3D trajectories for both tissue and instruments.}

\change{Despite these strengths, several limitations remain. Since the annotations are generated from monocular videos, the recovered geometry is sequence-consistent but pseudo-scale, which is suitable for relative 3D motion analysis but requires additional calibration for metric robot-workspace reasoning. In addition, the quality may decrease under particularly challenging visual conditions, such as long-term instrument disappearance, severe occlusion, fast motion. Future work can further improve metric accuracy and annotation reliability by incorporating robot kinematics, stereo calibration, tool geometry, anatomical priors, and uncertainty-aware filtering.}

\change{Overall, SurgTAP-3D demonstrates that large-scale contact-aware 3D motion annotation from in-vivo surgical videos is practical and scalable, providing a useful data resource and construction pipeline for spatially aware surgical perception.}

\section{Conclusion}
In this work, we present SurgTAP-3D, the first unified framework and large-scale in-vivo dataset for contact-aware 3D surgical point tracking from uncalibrated monocular videos. By integrating 3D foundation priors with \change{dynamic} bundle adjustment and global 4D optimization, our pipeline recovers spatially coherent, temporally smooth, and physically plausible 3D trajectories in both camera and \change{global coordinates}. Built upon 4,549 action clips and over a million 2D/3D trajectories across diverse procedures, SurgTAP-3D substantially advances the scale, quality, and realism of publicly available surgical motion data. Comprehensive benchmarks and downstream applications further demonstrate its effectiveness in improving 3D tracking, reconstruction, and instrument motion analysis. We believe this work establishes a scalable pathway for transforming raw surgical videos into structured 4D supervision, paving the way toward data-driven, spatially grounded, and generalizable autonomous surgical systems.

\bibliographystyle{IEEEtran}
\bibliography{IEEEabrv,revised_references}
\vfill

\end{document}